\documentclass[10pt,journal,compsoc]{IEEEtran}
\usepackage{amsmath,amsfonts}
\usepackage{algorithmic}
\usepackage{algorithm}
\usepackage{array}
\usepackage[caption=false,font=normalsize,labelfont=sf,textfont=sf]{subfig}
\usepackage{textcomp}
\usepackage{stfloats}
\usepackage{url}
\usepackage{verbatim}
\usepackage{graphicx}
\usepackage{cite}
\usepackage{cuted}
\usepackage{caption}

\usepackage{multirow}
\usepackage{color,xcolor}
\usepackage{ragged2e}
\usepackage{hyperref}

\begin{document}

\title{SpeechAct: Towards Generating Whole-body Motion from Speech}

\author{Jinsong Zhang$^{\dagger}$, Minjie Zhu$^{\dagger}$, Yuxiang Zhang, Zerong Zheng, Yebin Liu,~\IEEEmembership{Member,~IEEE}, \\ Kun Li$^{*}$,~\IEEEmembership{Senior Member,~IEEE} 

\IEEEcompsocitemizethanks{

\IEEEcompsocthanksitem  $\dagger$ Equal contribution.
\IEEEcompsocthanksitem $^{*}$ Corresponding author: Kun Li (Email: lik@tju.edu.cn)
\IEEEcompsocthanksitem Jinsong Zhang, Minjie Zhu, and Kun Li are with the College of Intelligence and Computing, Tianjin University,
Tianjin 300350, China.

\IEEEcompsocthanksitem Yuxiang Zhang and Yebin Liu are with the Department of Automation, Tsinghua University, Beijing 100084, China.

\IEEEcompsocthanksitem Zerong Zheng is with NNKosmos Technology, Hangzhou, P.R.China.
\protect\\}
}

\markboth{Journal of \LaTeX\ Class Files,~Vol.~14, No.~8, August~2021}%
{Zhang \MakeLowercase{\textit{et al.}}: SpeechAct: Towards Generating Whole-body Motion from Speech Audio}



\IEEEtitleabstractindextext{
\begin{center}\setcounter{figure}{0}
    \includegraphics[width=.9\textwidth]{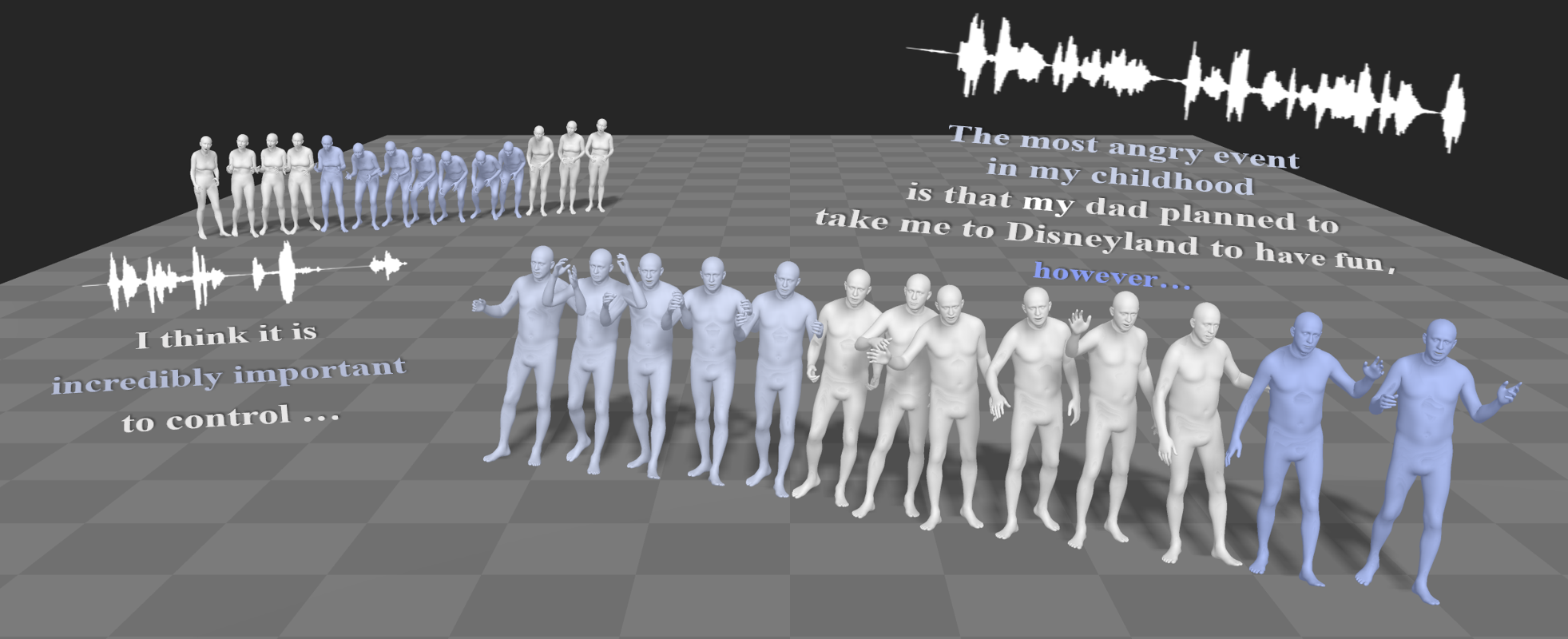}
    \captionsetup{font={scriptsize}}
    \captionof{figure}{Given an audio input, our model can generate {natural and diverse} human motion sequences. There are two samples that are uniformly sampled from generated space. The human body meshes corresponding to the same text color indicate the motion generated by the driven speech content.}
    \label{fig_teaser}
  \end{center}
\justifying
\begin{abstract}
Whole-body motion generation from {speech audio} is crucial for computer graphics and immersive VR/AR.
Prior methods struggle to produce natural and diverse whole-body motions from speech.
In this paper, we introduce a novel method, named SpeechAct, based on a hybrid point representation and contrastive motion learning to boost realism and diversity in motion generation. Our hybrid point representation leverages the advantages of keypoint representation and surface points of 3D body model, which is easy to learn and helps to achieve smooth and natural motion generation from speech audio. We design a VQ-VAE to learn a motion codebook using our hybrid presentation, and then regress the motion from the input audio using a translation model. To boost diversity in motion generation, we propose a contrastive motion learning method according to the intuitive idea that the generated motion should be different from the motions of other audios and other speakers. We collect negative samples from other audio inputs and other speakers using our translation model. With these negative samples, we pull the current motion away from them using a contrastive loss to produce more distinctive representations. In addition, we compose a face generator to generate deterministic face motion due to the strong connection between the face movements and the {speech audio}. Experimental results validate the superior performance of our model. 
The code is available at \href{http://cic.tju.edu.cn/faculty/likun/projects/SpeechAct/index.html}{http://cic.tju.edu.cn/faculty/likun/projects/SpeechAct/index.html}
\end{abstract}

}

\maketitle
\IEEEdisplaynontitleabstractindextext

%
\IEEEpeerreviewmaketitle

\section{Introduction}
\IEEEPARstart{H}{uman} { motion generation from {speech audio~\cite{yi2023generating,yang2023diffusestylegesture,habibie2021learning}}  
is a critical area in computer graphics and immersive VR/AR \cite{le2012live,aristidou2022rhythm,li2023audio2gestures,wang2019combining}, which has been extensively studied as a way of human behavior understanding \cite{yi2023generating}.}
Given a speech recording, {the goal is to generate a spectrum of diverse yet natural motion sequences,} which is in line with real-life scenarios and can meet the varying needs of different individuals.

Existing approaches concentrate on translating speech to the motion of the partial body \cite{ginosar2019gestures,habibie2021learning, ao2022rhythmic, Guo_2022_CVPR}. They use keypoints of the body as their motion representation, which is easy to learn and contains local details for hands. However, the keypoints leave degrees of freedom undefined, \emph{e.g.}, the rotation of a limb, leading to inaccurate and unrealistic results when fitting or animating a full 3D body.
For the whole-body motion generation, TalkShow \cite{yi2023generating} adopts the parametric representation of SMPL-X body mesh as the motion representation, and proposes a two-stage model to recover the full 3D body.
However, we observe that the motion generated by TalkShow often contains discontinuous results, for example, jitters and disharmony contact between the feet and the ground.
This is because they use parametric representation defined in the axis-angle space using the kinematic tree, which may introduce {the discontinuity caused by the axis-angle representation, leading to the difficulty of modeling continuous latent space} \cite{kolotouros2019convolutional,  tian2023recovering, zhang2021we}.
Moreover, due to the inherent one-to-one mapping {in our real life and training pairs,}
all the above approaches tend to learn an average motion lacking diversity.
However, in fact, given a speech recording, different speakers tend to exhibit varying motions in different situations.
Therefore, how to generate diverse results is also a challenging problem.

In this work, to generate smooth and natural motion, we propose a hybrid point representation for whole-body motion generation from {speech audio}.
Specifically, the keypoint representation is easy to learn and contains local details for hands, but lacks surface constraints to obtain the 3D body model.
To combat this problem, we adopt the surface points of SMPL-X body mesh as a part of our hybrid point representation to eliminate the ambiguity due to the keypoint representation.
Therefore, defined in Euclidean space, our hybrid point representation is easy to learn and encompasses global constraints and local details for whole-body motion generation. This helps generate smooth and natural results, \emph{e.g.}, avoiding foot skating.
Besides, to provide an easy-to-use output representation that can be used in many applications, \emph{i.e.}, SMPL-X body mesh, we design a generator to recover the parameters of SMPL-X from our hybrid representation.

{With this hybrid point representation, we can generate smooth and natural results,}
however, similar to previous approaches, the diversity of generated results is still limited due to the one-to-one mapping in the training data.
{This one-to-one mapping in the training data makes the model capture similar motions, \emph{e.g.}, certain habitual body motions of a speaker and similar body motions of different speakers due to specific meanings, but this limits the model to generate diverse motions.}
To boost the diversity of generated results, an intuitive idea is that, the generated motion of a specific audio should be different from the generated motions of other audios and other speakers.
To achieve this, we introduce a novel contrastive motion learning method to obtain distinctive motion representations.
Specifically, we first collect negative samples that are discrete motion representations from other audio inputs and other speakers using the translation model.
Then, we pull the current generated motion away from the negative samples by a contrastive loss, and thus the generated motion can be more distinctive.
{By using the target motion as a positive sample, our approach encourages diverse motion learning while capturing similar motions.}

{Built on our hybrid point representation and motion contrastive learning method, we present a novel model, named SpeechAct, to generate natural and diverse body motion. This model first constructs a quantized motion codebook based on our presentation, and then regresses the distinct motion from {speech audio} using our contrastive motion learning.}
{In addition, to support whole-body motion synthesis, we compose a face generator to generate deterministic face motion due to the strong connection between the face movements and the {speech audio}. }
Experimental results compared with several state-of-the-art methods validate the effectiveness of our model.
{We give an application that animates avatars using speech inputs, which can be applied in immersive AR/VR \cite{li2023high,zhao2022high}.  Moreover, our model can generate promising results conditioned on different languages.}
Fig. \ref{fig_teaser} presents two samples generated by our model. The human body meshes corresponding to the same text color indicate the motion generated by the driven speech content. It can be seen that our approach not only generates smooth sequences of movements containing diverse poses, but also ensures that these sequences correspond to the audio content.

Our main contributions can be summarized as follows:
\begin{itemize}
    \item {We design SpeechAct, a novel framework based on a new representation to generate whole-body motion from speech audio, which can produce more natural and diverse results.}
    \item We introduce a novel hybrid point representation that contains global constraints and local details for whole-body, and a generator that can recover parameters of SMPL-X body mesh, which can be used for avatar animation.
    \item We propose a contrastive motion learning method to learn a more distinctive motion representation, which improves the ability of our model to generate diverse results.
    {\item Experimental results demonstrate our model can generate natural and diverse results. Our method can also be generalized to other languages.  }
    
\end{itemize}

\section{Related Work}
\label{sec:related}

{\subsection{Human Motion Generation}}
{Human motion generation is the task of synthesizing realistic and natural human movements for various applications, such as animation, gaming, robotics, and VR/AR. The data representations can be divided into two kinds, \emph{i.e.}, keypoint-based and rotation-based. The keypoint-based representation \cite{habibie2021learning,degardin2022generative,cao2020long} adopts a set of keypoints that are specific points on the body, \emph{e.g.}, joints. The keypoints are defined using 2D/3D coordinates in the pixel or world coordinate system. Some works \cite{zhang2021we, ma20233d} use sparse markers to represent human pose, which can provide more comprehensive information for rotation. However, the keypoint-based methods can provide local details for hands, but can not constrain the global shape, such as the rotation of limbs, which can lead to unrealistic results when fitting or animating a full 3D body.  The rotation-based methods \cite{zhang2022couch,hassan2021stochastic} adopt the rotation of the body part relative to their parent, which helps to constrain the body pose by modeling the body, face and hands jointly. However, this representation can generate discontinuous results due to its specific space \cite{kolotouros2019convolutional,  tian2023recovering, zhang2021we,li2023mili}, \emph{e.g.}, axis-angle space for SMPL-X body model.}

{In this paper, we propose a hybrid point representation for whole-body motion generation. Our hybrid point representation is defined in Euclidean space and is easy to learn, which encompasses global constraints and local details.}

\subsection{Whole-body Motion Generation from Speech}
Previous approaches focus on generating different parts of whole-body from speech, \emph{e.g.}, face, hands and body.
Existing speech-driven 3D face animation methods \cite{cudeiro2019capture, fan2022faceformer} rely on 4D face scan datasets to train their models.
For body motion generation, two kinds of methods have been explored: rule-based methods and learning-based methods.
Rule-based methods \cite{cassell2001beat, kopp2004synthesizing, levine2010gesture} convert the input speech to motions from a pre-collected motion database according to manually designed rules.
These methods are interpretable and controllable, but are difficult to generate complex and realistic motions.
Due to limitations in existing datasets \cite{habibie2021learning,ginosar2019gestures,takeuchi2017creating,yoon2019robots}, previous learning-based methods primarily concentrate on generating partial human body motions from speech \cite{habibie2021learning, ahuja2020no, bhattacharya2021speech2affectivegestures, kucherenko2019analyzing, kucherenko2020gesticulator}.
Most of them adopt keypoints as motion representation, achieving accurate results closely approximating the ground-truth keypoints but leading to inaccurate and unrealistic results when fitting or animating a full 3D body.
Besides, these methods generate deterministic results. This means that they fail to generate diverse results given the same audio signal, which is inconsistent with cognitive understanding.

Some approaches cooperate GANs \cite{ahuja2020style,liu2022learning,yoon2020speech,wang2019combining}, VAEs \cite{li2021audio2gestures, li2023audio2gestures, qian2021speech}, VQ-VAEs \cite{ao2022rhythmic,yazdian2022gesture2vec}, and normalizing flows \cite{alexanderson2020style, yang2023keyframe} to increase the diversity of the generated results, but the results are inadequate \cite{yi2023generating}.
{Recently, due to the great success of diffusion models \cite{ho2020denoising}, some methods \cite{ao2023gesturediffuclip,alexanderson2023listen,yang2023diffusestylegesture, li2024lodge} introduce diffusion models to generate human motion from speech, which can generate promising results. However, this kind of method requires a large number of diffusion steps, and thus has low generation efficiency.}
Besides, modeling part of the whole-body is insufficient for a comprehensive understanding of human behavior. 
To generate whole-body motion, {most similar to our work,}
Yi \emph{et al.} \cite{yi2023generating} fit the SMPL-X body mesh based on the speaker-specific gesture dataset \cite{ginosar2019gestures}, but the motions in their dataset focus on face and upper-body gestures.
Based on their dataset, they propose to generate diverse motions for the body and hands using two independent codebooks. 
{Liu \emph{et al.} \cite{liu2023emage} propose a new whole-body co-speech dataset, \emph{i.e.}, BETAX dataset, and design a masked audio gesture transformer to regress the gesture from the audio input. }
However, because of the inherent one-to-one mapping between input and output, the model struggles to generate natural and diverse results.
Besides, they adopt parameters of SMPL-X body mesh as the motion representation, which introduces complexity for the prediction task \cite{mahendran2018mixed} and can lead to discontinuous problems.

In this paper, we introduce a novel hybrid representation by combining keypoints and surface points of SMPL-X body mesh, which helps to generate continuous results and is easy to expand to other applications.
Besides, we introduce a novel contrastive motion learning method by collecting negative motion samples and pulling the generated motion away from them, which promotes the model to generate more diverse results.

\subsection{Contrastive Learning}
{Contrastive learning is a powerful self-supervised learning method, which has achieved great progress in many computer vision and computer graphic tasks. Traditionally, the central idea is to take a sample as an anchor, compare it with other samples, and pull it closer to positive samples while pushing it apart from negative samples. By doing so, we can get a meaningful representation. Though designed for 2D image tasks \cite{Jiang_2023_CVPR,park2020contrastive}, contrastive learning has been used in 3D tasks \cite{10091230,sanghi2020info3d} and multi-modal tasks \cite{ao2023gesturediffuclip,zhao2023taming}. Ao \emph{et al.} \cite{ao2023gesturediffuclip} adopt contrastive learning to learn a gesture-transcript joint embedding space to align the gesture space and the transcript space. Zhao \emph{et al.} \cite{zhao2023taming} use a similar idea to achieve music-driven dance generation.}

{Different from previous methods, instead of aligning the multi-modal information to obtain a joint embedding space, we propose a contrastive motion learning method to boost the diversity of generated results.}

\begin{figure*}[!ht]
		\centering
		\includegraphics[scale=0.6]{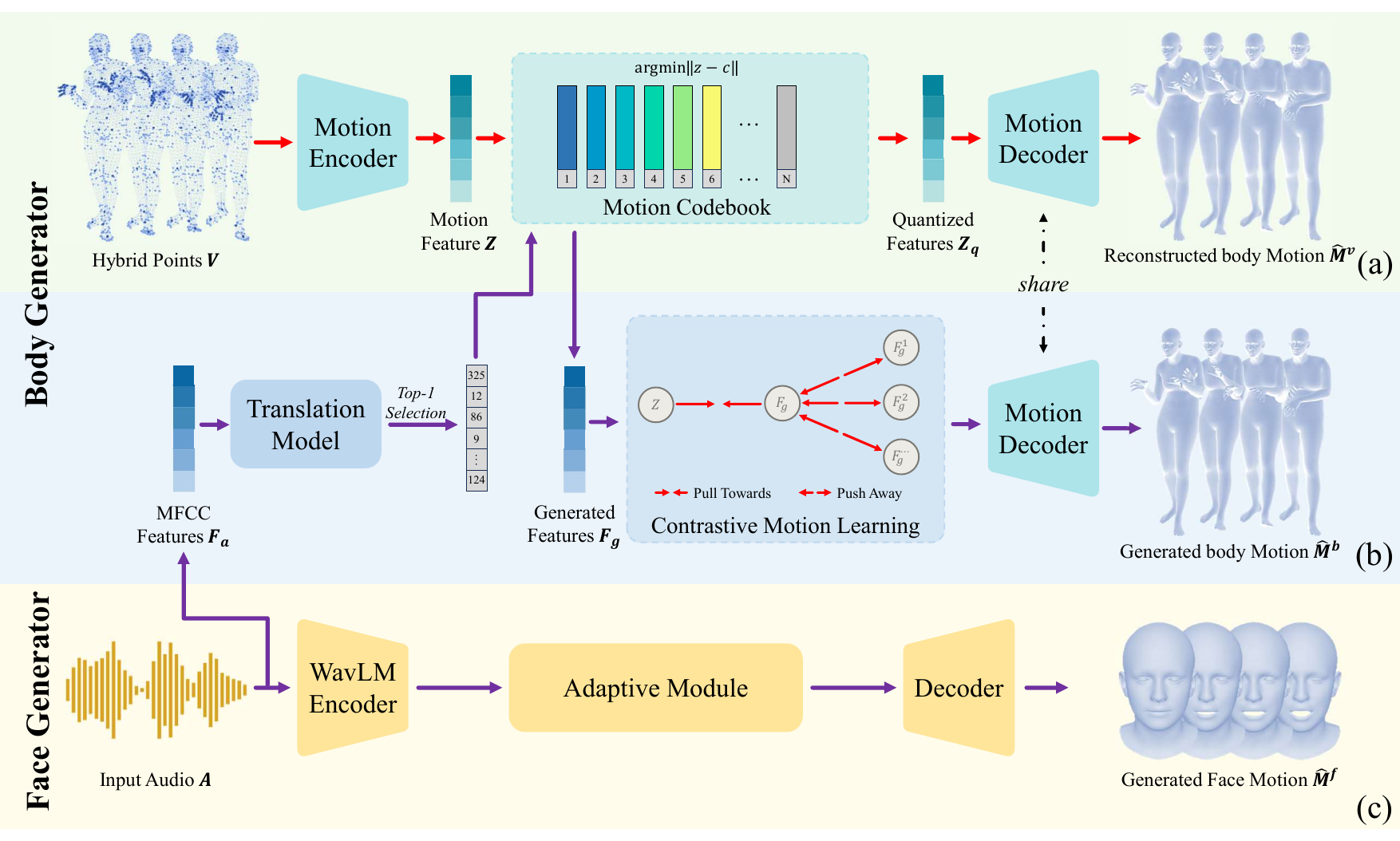}
		\caption{The detailed architecture of SpeechAct. To generate whole-body motion, our model includes a two-stage body generator to generate diverse motions for the body and hands and a face generator to output deterministic results. Specifically, our model includes: (a) a VQ-VAE based on our proposed hybrid point representation to learn a motion codebook, (b) a translation model with a contrastive motion learning method to generate diverse motion codes from the learned motion codebook, and (c) an encoder-decoder architecture to generate deterministic face motion. {The red lines indicate these modules are used for training, and the purple lines mean that these modules are applied for both training and inference.}}
		\label{fig_net}
    \vspace{-0.3cm}
	\end{figure*}
 
\section{Method}
\label{method}
 
Given a speech recording, our work aims to generate diverse whole-body motion sequences that are in harmony with the provided speech.
We design a novel framework, named SpeechAct, to achieve this.
The most significant differences with existing approaches are that, to generate natural and diverse results for the body and hands, we propose a hybrid point representation to form our motion efficiently, and a contrastive motion learning method to distinguish the generated motion from other motions to boost the diversity of the generated results.
Fig. \ref{fig_net} shows the overall framework of our SpeechAct.
Our model consists of two generators: a body generator (Sec. \ref{bgen}) and a face generator (Sec. \ref{fgen}).
Because the audio signal is not very closely related to hand and body gestures \cite{yi2023generating}, the body generator aims to generate natural and diverse results.
For an effective and versatile motion representation, we introduce a hybrid point representation (Sec. \ref{hrep}) and develop a generator to obtain SMPL-X model parameters. 
To ensure the generation of coherent and diverse results, we leverage advancements in VQ-VAE to learn a quantized motion space (Sec. \ref{vqvae}). Building upon this quantized motion space, we propose a translation model (Sec. \ref{regress}) and introduce a novel contrastive motion learning method, allowing us to transform audio signals into motion representations while promoting diversity in body and hand motions.
In contrast, {speech content, especially phonetic information, is closely related to face motion, notably lip motion}, thus the face generator is designed using an encoder-decoder architecture and is responsible for generating deterministic results.

\subsection{Body Generator}
\label{bgen}

For the body generator, we aim to generate natural and diverse motion sequences for the body and hands given an audio input. 
Previous approaches \cite{ginosar2019gestures, habibie2021learning} adopt keypoints for motion representation, limiting their ability to fully reconstruct or animate a 3D body.
TalkShow \cite{yi2023generating} utilizes SMPL-X model parameters, resulting in discontinuous motion sequences. 
To generate more realistic and continuous results, we propose a hybrid point representation with a generator, which can generate continuous results efficiently and deliver the SMPL-X model accurately.
Besides, because the input and the output are one-to-one mappings, previous methods struggle to generate natural yet diverse results due to the challenges of balancing constraints between precision and diversity.
To alleviate this problem, we design a two-stage model and introduce a novel contrastive motion learning method to boost the diversity of generated results.
Fig. \ref{fig_net} shows the framework of our body generator.
Specifically, {we first learn a motion codebook by encoding motion sequences into quantized codes. To synthesize more natural hand motion, we model the body and hands separately using three different codebooks.}
Subsequently, our translation model autoregressively transforms audio signals into motion sequences. 
With the contrastive motion learning method, the translation model can generate more distinctive motion representations, which helps to generate diverse results.

\subsubsection{Hybrid Point Representation}
\label{hrep}

\begin{figure}[!ht]
		\centering
		\includegraphics[scale=0.35]{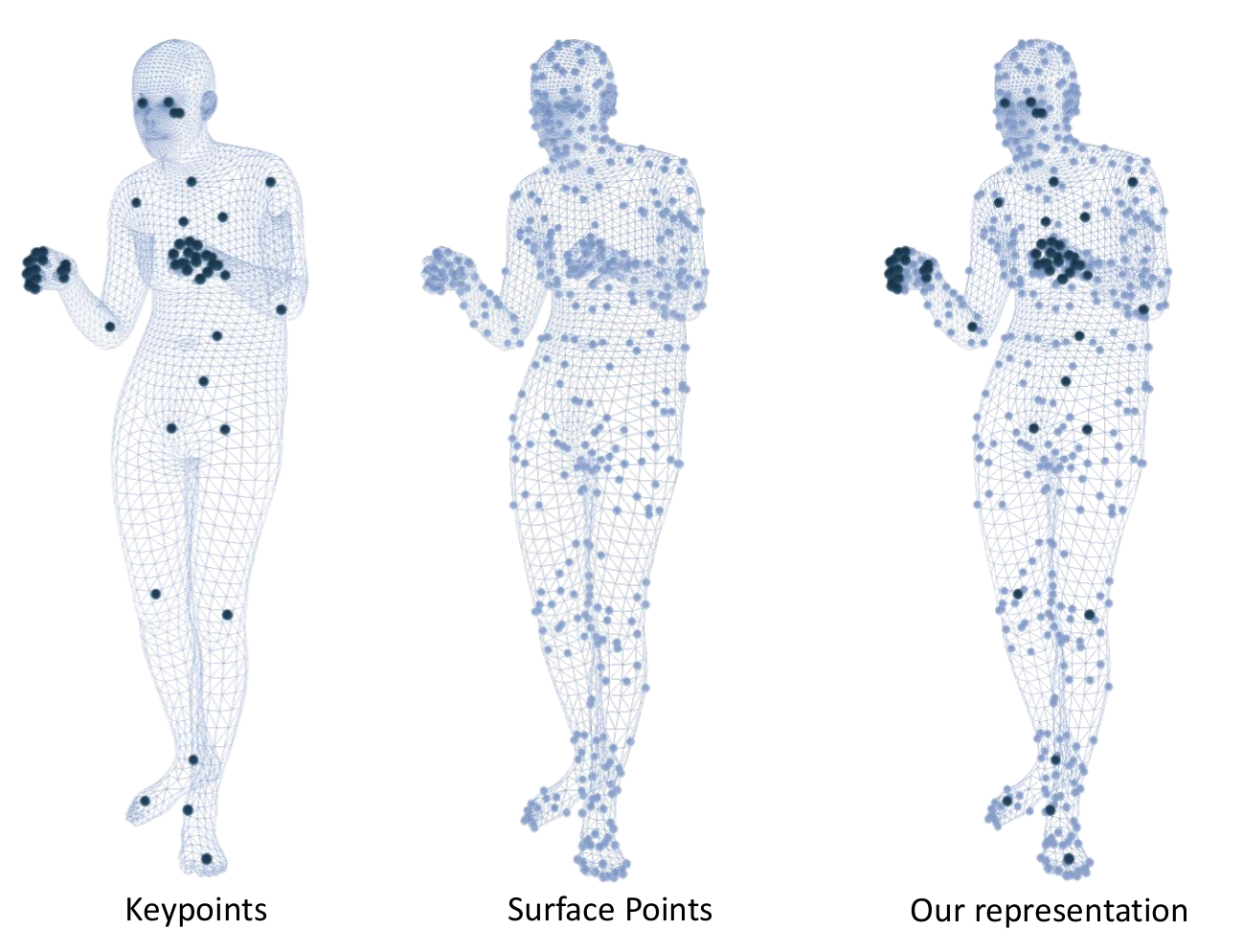}
		\caption{{The overview of our hybrid representation}}
		\label{fig_rep}
        \vspace{-0.4cm}
    
	\end{figure}
 
{As shown in Fig. \ref{fig_rep},}
our hybrid points representation consists of two vital components: surface points $v_s$ (the points in light blue) and keypoints $v_k$ (the points in dark blue). Surface points are extracted from the SMPL-X body mesh 
{using mesh sampling operation \cite{coma},}
serving as a global constraint to the freedom of the body. 
{On the other hand, keypoints, which are defined by SMPL-X body mesh and can be obtained using the joint regressor from SMPL-X  \cite{pavlakos2019expressive}, capture fine-grained details, particularly in areas like the hands, enriching the representation.}
{In practice, we adopt all keypoints, \textit{i.e.}, 55 keypoints, of SMPL-X body mesh, and use 431 points sampled from the surface. As both keypoints and surface points are defined in Euclidean space, the final representation $v \in \mathcal{R}^{486\times 3}$}, results from the concatenation of $v_s$ and $v_k$, with each element representing a 3D location. 
This combined representation facilitates the capture of both global constraints and local details.
{This representation is employed to represent a motion sequence, denoted as $V = \{v_t\}_{t=0}^{T}$, where $T$ represents the number of frames in a motion sequence.} It encapsulates the dynamic evolution of the hybrid point-based representation over time, enabling comprehensive modeling and analysis of human body motion.

Inspired by \cite{kolotouros2019convolutional}, to eliminate the gap between the SMPL-X body mesh and our representation, {we design a simple generator $\mathcal{G}_P$, where the input is $v_t$ and the output is the parameters of the SMPL-X body mesh.}
This generator $\mathcal{G}_P$ consists of a series of neural network layers, specifically a stack of three residual blocks, each with varying input and output channels. These layers are responsible for transforming the input point representation into a more meaningful intermediate feature representation. Subsequently, a final convolutional layer is employed to produce the desired output, which represents the parameters of the SMPL-X body mesh. {By doing so, even training the body and face separately, our model can ensure consistency between body and face to generate seamless and continuous results.}
This generator is trained with our body generator, which is represented in Sec. \ref{vqvae}.

\noindent
\textbf{Discussion.}
Our hybrid point representation has two advantages compared with previous approaches:
\begin{itemize}
    \item Compared to the keypoint representation, our hybrid representation can eliminate the ambiguities when recovering the shape and pose of the whole-body. Besides, with the generator $\mathcal{G}_P$, our representation can be transformed to SMPL-X body mesh easily, and this process is differentiable.
    \item Our representation is defined in the Euclidean space. This choice not only facilitates the generation of continuous motion, \emph{e.g.}, avoiding foot skating, but also enhances the model's learning process compared to parameter-based representations.
\end{itemize}

\subsubsection{Quantized Motion Modeling}
\label{vqvae}

As shown in Fig. \ref{fig_net} (a), the first stage of our body generator is quantized motion modeling.
{Given a set of motion sequences $\mathcal{M}=\{M_k\}_{k=0}^K$, where $K$ is the number of motion sequences, we first extract our hybrid point representation $\mathcal{V}=\{V_k\}_{k=0}^K$.
With a sample $V$ from $\mathcal{V}$, we adopt a Vector Quantized-VAE (VQ-VAE) \cite{van2017neural} to learn a meaningful and compact motion space.
We first obtain motion features $Z\in \mathcal{R}^{\frac{T}{w_m}\times n_z}$ from the input motion sequence, where  $T$ is the number of frames in $V$, $n_z$ is the channel dimension of quantized features and $w_m$ is the temporal window size. }

Denote the learneable motion codebook as $C = \{ c_i\}_{i=1}^{N}$, where $N$ is the length of the codebook.
Then, we quantize the motion features by replacing them with the nearest codes in the codebook.
{Specifically, for each row $z_t$ in $Z$, the quantized feature can be obtained by:}
\begin{equation}
\label{eq:1}
    c_t = \arg \min \limits_{c_i\in C} ||z_t - c_i||_2.
\end{equation}
Therefore, the quantized features $Z_q=\{c_t\}_{t=0}^{\frac{T}{w_m}}$.
Afterward, the reconstructed motion $\hat{V}$ is delivered by a decoder.
To output the SMPL-X body mesh, we cooperate with the generator $\mathcal{G}_P$ in the motion decoder, which can transform our hybrid point representation into SMPL-X body mesh.

We train this VQ-VAE with the reconstruction loss and the vector quantization loss to obtain a meaningful motion space.
The reconstruction loss is used to recover the motion sequence accurately.
Following \cite{siyao2022bailando}, we recover the locations, velocities, and accelerations of the movements, mathematically:
\begin{equation}
\label{eq:2}
    \mathcal{L}_{rec} = ||\hat{V}-V||_1 + \alpha_1||\hat{V}^{'}-V^{'}||_1 + \alpha_2||\hat{V}^{''}-V^{''}||_1,
\end{equation}
where $V^{'}$ and $V^{''}$ are the first-order and second-order partial derivatives of motion representations $V$, and $\alpha_1$ and $\alpha_2$ are the balancing weights of the corresponding items.

The vector quantization loss can be written as:
\begin{equation}
\label{eq:3}
    \mathcal{L}_{vq} = ||\rm{sg}[Z]-Z_q||_2 + \beta ||Z-\rm{sg}[Z_q]||_2,
\end{equation}
where $\rm{sg}[ \cdot ]$ is the stop gradient function \cite{chen2021exploring}, the first item is the codebook loss and the second item is the commitment loss with the weight $\beta$. {The codebook loss and the commitment loss are used to align the vector space of the codebook and the outputs of the encoder. The codebook loss lets the quantized features from the codebook close to the outputs of the encoder, and the commitment loss brings the outputs of the encoder close to the codes in the codebook.}

{In practice, we adopt three codebooks, \textit{i.e.}, left-hand codebook, right-hand codebook, and body codebook, to model more detailed motions for the body and hands, respectively. Therefore, the encoded feature $Z$ can be split into three parts $Z^{lh}, Z^{rh}, Z^{b}$ according to different body parts of our hybrid representation before quantization, and  these three features can be quantized using Eq. \ref{eq:1}. The quantized feature $Z_q$ is the concatenation of left-hand quantized feature $Z_q^{lh}$, right-hand quantized feature $Z_q^{rh}$, and body quantized feature $Z_q^{b}$. Therefore, though we adopt three codebooks, they can be optimized using Eq. \ref{eq:2} and Eq. \ref{eq:3} simultaneously.}

To train the generator $\mathcal{G}_P$ simultaneously, we adopt the L1 distance to regress the parameters of SMPL-X body mesh, mathematically:
\begin{equation}
    \mathcal{L}_{sx} = ||\mathcal{G}_P(\hat{V}) - \theta||_1,
\end{equation}
where $\theta$ is the ground-truth parameters of SMPL-X body mesh.

\subsubsection{Translation Model}
\label{regress}
With the learned motion codebook, as shown in Fig. \ref{fig_net} (b), we aim to translate the audio signal into the motion representation from the motion codebook, and then output the motion sequence using the learned motion decoder in the second stage.
{Speech content, especially phonetic information, is closely related to face motion, notably lip motion, while its link to body/hand motion is weaker. Compared with phonetic information in the audio signal, speech rhythm and beat are more related to body and hand motion. Therefore,}
given the input audio signal $A = \{a_t\}_{t=0}^T$, we first transform it to 64-dimensional Mel-Frequency Cepstral Coefficients (MFCC) features $F_a$ containing rhythm and beat information. 
To regress the motion representation with speaker information, we design a translation model $\mathcal{T}$ to translate the input audio signal into the motion codes.
Specifically, we concatenate the speaker embedding, \emph{i.e.}, a one-shot embedding, with $F_a$, and employ several retention blocks \cite{sun2023retentive} to extract the regressed features $F_r \in \mathcal{R}^{\frac{T}{w_m}\times N}$, where $w_m$ is the temporal window size and $N$ is the length of the motion codebook.
Subsequently, we adopt a SoftMax layer to obtain the regressed logits, and select the highest score at each temporal window as the index of the quantized feature.
Thus, we can obtain the generated features $F_g$ by selecting features according to the selected indexes from the learned motion codebook.
With the learned motion decoder, we can obtain natural results in harmony with the input audio. 
To regress natural results, we adopt cross-entropy loss $\mathcal{L}_{ce}$ to optimize the translation model.
{As an auto-regressive model, the translation model can generate motions of arbitrary length by adjusting the input sequence of previously predicted tokens.} {Besides, during inference, we randomly initialize the first token using the index of the codebook to generate diverse results.}

\begin{figure}[!t]
		\centering
		\includegraphics[scale=0.2]{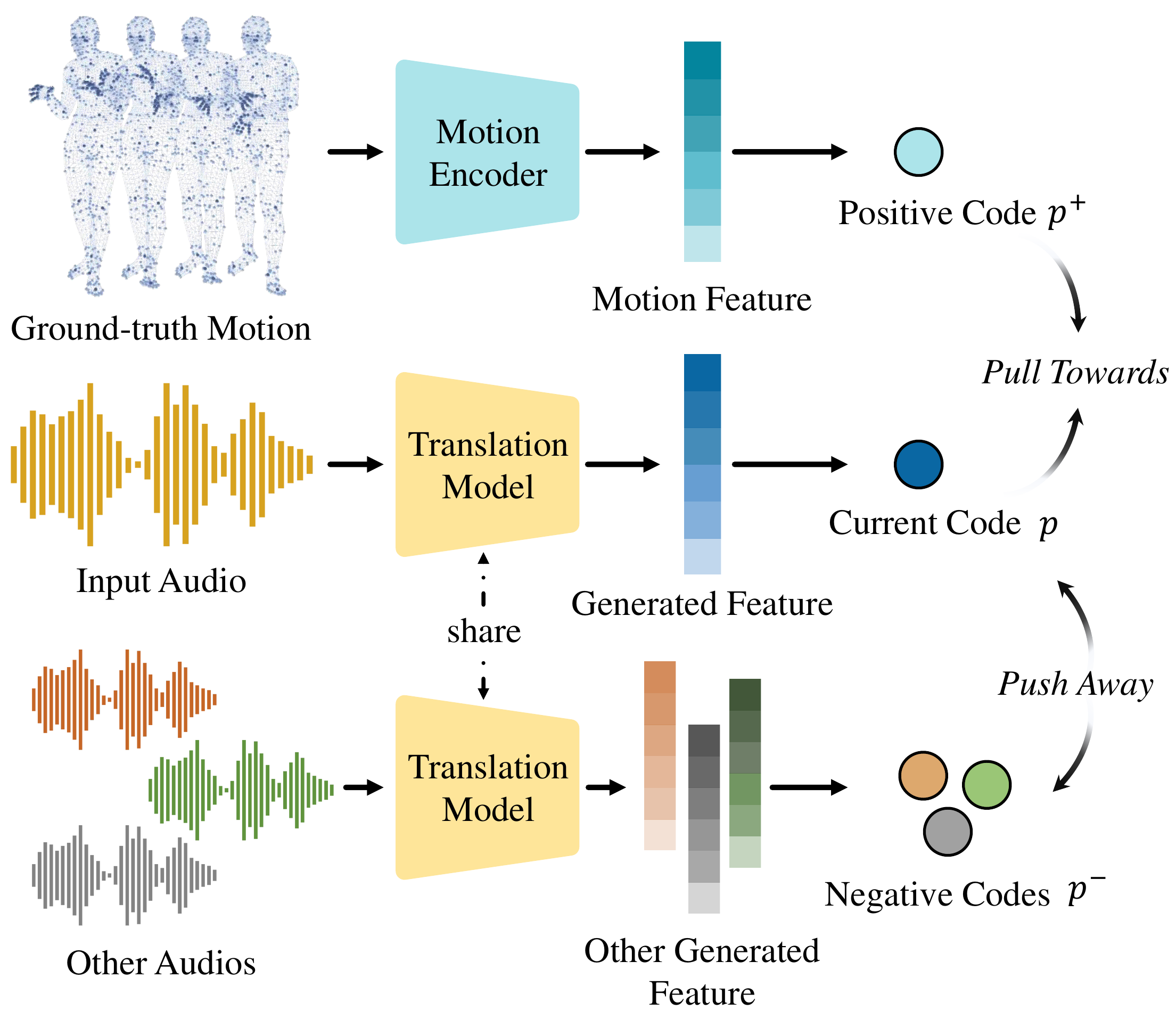}
		\caption{Details of contrastive motion learning. We take the quantized features from the ground-truth motion as the positive sample, and the generated features from other audios as the negative samples. By pulling away the current generated feature from the negative samples, we can obtain more distinctive representations. }
		\label{fig_net_con}
	\end{figure}
 
\noindent
\textbf{Contrastive Motion Learning.}
Because the input audio and the output motion are one-to-one mappings {in training pairs}, the diversity of the generated results is still limited.
An intuitive idea is that, given a speech recording, the motion for a specific speaker can be different from other speakers, and can be different from the motions driven by other speech recordings.
To achieve this, we introduce a novel contrastive motion learning method, which can distinguish the current generated motion from other motions driven by different audios or generated by other speakers.
As shown in Fig. \ref{fig_net_con}, different from previous works \cite{ao2023gesturediffuclip, ao2022rhythmic} using contrastive loss to align multi-modal information, \emph{e.g.}, text and motion, we take the early generated quantized features from other speakers and other speech recordings as the negative samples, and the ground-truth quantized features as the positive sample to boost the diversity of generated results.
Specifically, we adopt global average pooling along time dimension to obtain motion codes as the negative or positive samples.
Evidently, the positive sample is high-quality and in harmony with the input audio signal, while the negative samples are low-quality and have no connection with the input audio signal.
Let $p$ denote the current generated features $F_g$, $p^+$ that of groud-truth quantized feature from $V$, and $\{ p_t^-\}_{t=1}^{L}$ the generated features from $L$ negative samples.
To boost the diversity of the generated results, our goal is to ensure that the generated features are different from the negative samples, \emph{i.e.}, minimizing mutual information.
Therefore, the contrastive loss can be formulated as:
\begin{equation}
    \mathcal{L}_{cm} = - \log \frac{\exp (\frac{p^T p^+}{\tau})}{\exp (\frac{p^T p^+}{\tau}) + \sum_{j=1}^{L} \exp (\frac{p^T p_j^-}{\tau})},
\end{equation}
where $\tau$ is set to 0.7 and $L$ is set to 1024  in our experiments.
{By doing so, our model can produce more realistic and diverse motions while maintaining fidelity to ground-truth motions.}

\subsection{Face Generator}
\label{fgen}

{In this work,  our goal is to generate whole-body motion from speech audio, and the most important part is body generation due to its diverse character.  To support the overall whole-body motion synthesis, we include a face generator to generate facial motion from speech audio.}
Because the face is closely related to the input signal, {especially phonetic information}, as shown in Fig. \ref{fig_net} (c), we adopt an encoder-decoder architecture to regress the facial movement.
Specifically,  
we first encode the audio signal  $A = \{ a_i\}_{i=0}^T$ using a pre-trained {WavLM model \cite{chen2022wavlm}}, where the feature extraction is adapted to generate a representation suitable for downstream tasks.
This representation is concatenated with speaker embedding to obtain speaker information, and then refined by an adaptive module to extract and aggregate the audio features adaptively. 
{The adaptive module consists of a residual block, followed by an attention block, and then two more residual blocks. This hierarchical arrangement of convolutional and attention mechanisms aims to robustly capture and understand the speaker-specific characteristics and the speech content to generate more reliable results. }
Subsequently, a series of residual blocks with attention mechanisms are used to decode the audio representation, transforming it into a deterministic face motion sequence. The detailed architecture can be found in the supplemental document.
{It should be noted that the face generator serves as a complementary aspect and can be replaced by state-of-the-art facial motion synthesis models \cite{ma2024diffspeaker,zhao2024media2face,sun2024diffposetalk,stan2023facediffuser}.}
The face generator is trained with face reconstruction loss: L1 distance for expressions and L1 distance {for the first order of the face motion representations.}

\subsection{Implemention Details}
We conduct our experiments on a desktop with a GeForce RTX 3090 GPU.
For the face generator, we adopt the parametric representation for deterministic results, and the parameters of WavLM are frozen.
For the body generator, we adopt our proposed hybrid points representation.
We first train the VQ-VAE to learn the motion codebook, and the generator from our representations to obtain parameters of SMPL-X.
{The weights ${\alpha_1, \alpha_2, \beta}$ of different losses are set to $0.5, 0.5, 0.25$. We set the channel dimensions of quantized features $Z_q^{lh}, Z_q^{rh}, Z_q^{b}$ as $128, 128, 512$. The length of the temporal window size is set to $\frac{1}{30}$ s for each input audio. }
Then, we train the translation model and freeze the parameters of the motion decoder.
For all training models, we adopt the Adam \cite{kingma2014adam} optimizer with coefficients 0.9 and 0.999 for computing running averages of gradients and their squares.
The learning rates for different stages are all set to 1e-4.

\section{Experimental Results}

\label{exp}

\subsection{Experimental Setting}

\noindent
\textbf{Dataset.}
To train and evaluate our model, we adopt BEAT2 dataset \cite{liu2023emage} as the benchmark to conduct experiments.
{The BEAT2 dataset contains 60 hours captured from 25 speakers in English, which is split into BEAT2-Standard (27 hours) and BEAT2-Additional (33 hours) based on the type of speech and conversation sections. We split BEAT2-Standard into  80\%, 10\%, and 10\% split for the train/val/test set.}

\noindent
\textbf{Baselines.}
To validate the performance of our model, {We first compare with TalkShow \cite{yi2023generating}, which is the most related work that generates whole-body motion from speech. Besides, we also compare with three state-of-the-art methods: Audio2Gesture \cite{li2021audio2gestures}, DiffStyle \cite{yang2023diffusestylegesture} and EMAGE \cite{liu2023emage}. Because Audio2Gesture and TalkShow are trained using other part body dataset, we re-train them on BEAT2 dataset using the public codes. Besides, we implement two baselines for generating body and hand motions to validate the effectiveness of our model. All the experiments are conducted on the same training and testing set. It should be noted that DiffStyle and  EMAGE are tested on 1 speaker. Therefore, we adopt the same speaker to evaluate their performance additionally for a fair comparison with them. }

\begin{itemize}
    
\item
\textbf{Audio Enc-Dec.}
Given a speech recording, this model first encodes the input and then decodes the features to deliver the motion, which is the same as \cite{ginosar2019gestures}.

\item
\textbf{Motion VAE.}
In this model, we first train a VAE using motion sequences to learn a latent space.
Then, given a {speech audio}, we encode the audio signal and concatenate it with a sampling latent code from latent space to output the generated motion using a decoder.





\end{itemize}

\noindent
\textbf{Metrics.}
For the body and hands, we aim to generate natural and diverse results in harmony with the input audio. 
Therefore, we adopt several evaluation metrics to validate the diversity and realism of generated results. 
\begin{itemize}
\item
\textbf{FID:}
{{We calculate Fr\'echet Inception Distances (FID) \cite{heusel2017gans} using kinetic features \cite{onuma2008fmdistance} (FID-k) and geometric features (FID-g) to measure the quality of generated results. Kinetic features are defined by motion velocities and energies, while geometric features are defined by joint angles and relative joint positions. }}

\item
\textbf{BeatAlign: }
As used in \cite{siyao2022bailando}, we employ chamfer distance between beats of the input audio and movements of the body and hands to measure the harmony between them, which indicates the quality of generated results.

\item
\textbf{Div-in: }
Following \cite{ng2022learning}, we evaluate the diversity of motions corresponding to an individual {speech audio} through variations in body poses across the temporal sequence, which is denoted as Div-in.

\item
\textbf{Div-out: }
Given a specific {speech audio}, diverse results for the same speaker are expected. To evaluate this, for all audio clips in our test set,
we calculate the average L2 distance between the two motion clips generated from an audio clip, denoted as Div-out. 

\item 
\textbf{FSR:} With our hybrid representation, our model can generate smooth results alleviating foot skating. To validate this character, following \cite{karunratanakul2023guided, li2024lodge},  we identify frames where the foot slides beyond a certain distance while in contact with the ground (i.e., foot height \textless 5 cm) as foot-skating frames. We report the Foot Skating Ratio (FSR), which measures the proportion of such frames. 

\end{itemize}

\begin{table}[!t]
	\renewcommand{\arraystretch}{1.3}
	\small%
 
	\setlength{\tabcolsep}{1.mm}
	\begin{center}
		\caption{{Quantitative comparison of body motion on BEAT2 dataset with several baselines. Ours$^{*}$ means we evaluate our method on the single speaker for a fair comparison with DiffStyle and EMAGE. Note that we report FSR $\times 10^{-2}$. }}
  \scalebox{0.83}{
		\begin{tabular}{c|cccccc}
			\hline
			{Model} &  FID-k $\downarrow$ &  FID-g $\downarrow$ & BeatAlign $\uparrow$  & Div-in $\uparrow$  & Div-out $\uparrow$ & {FSR} $\downarrow$  \\
			\hline
            Audio Enc-Dec  & \underline{10.36} & 8.907 & 0.5504 & 0.8632 & 0& 2.99 \\
            Motion VAE & 10.73 & 8.722 & 0.5363 & 0.8733 & 0.0571  & 6.78\\
			Audio2Gesture &20.43&4.826& \underline{0.5572} & 0.3893 & 1.312  & 1.26\\
           TalkShow  & 14.37&\underline{4.535}& 0.5479 & \underline{0.8966} & \underline{1.744} & \underline{0.61}\\
           Ours & \textbf{3.839} & \textbf{4.359} & \textbf{0.5603} & \textbf{0.9700} & \textbf{2.264} & \textbf{0.38}\\
   \hline \hline
            DiffStyle & 44.91 & \underline{6.576}  & \textbf{0.5811} & 1.539 & \underline{0.3340} & \underline{1.69} \\

            EMAGE  &\underline{6.620} &10.12 & 0.5487 & \underline{2.817} & 0 & 2.28\\
            Ours$^{*}$ & \textbf{5.979} & \textbf{5.505} & \underline{0.5704} & \textbf{3.335} & \textbf{7.840} & \textbf{1.19} \\
            \hline
		\end{tabular}}
  \label{tab_com}
  
\end{center}
\end{table}

 
            

\begin{table}[!t]
	\renewcommand{\arraystretch}{1.3}
	\small%
 
	\setlength{\tabcolsep}{4mm}
	\begin{center}
		\caption{{Quantitative comparison of face motion on BEAT2 dataset. We report LVD$\times 10^{-2}$ and LD $\times 10^{-2}$.}}
		\begin{tabular}{c|cc}
			\hline
			{Model} & LVD $\downarrow$ & LD $\downarrow$ \\
			\hline 
            TalkShow \cite{yi2023generating} & 4.158 & 0.8731\\
            EMAGE \cite{liu2023emage} & 4.359 & 0.9546 \\
            Ours & \textbf{3.829} & \textbf{0.7896} \\
            \hline
		\end{tabular}
  \label{tab_fcom}
  \vspace{-0.4cm}
\end{center}
\end{table}

\begin{figure*}[!t]
		\centering
		\includegraphics[width=\linewidth]{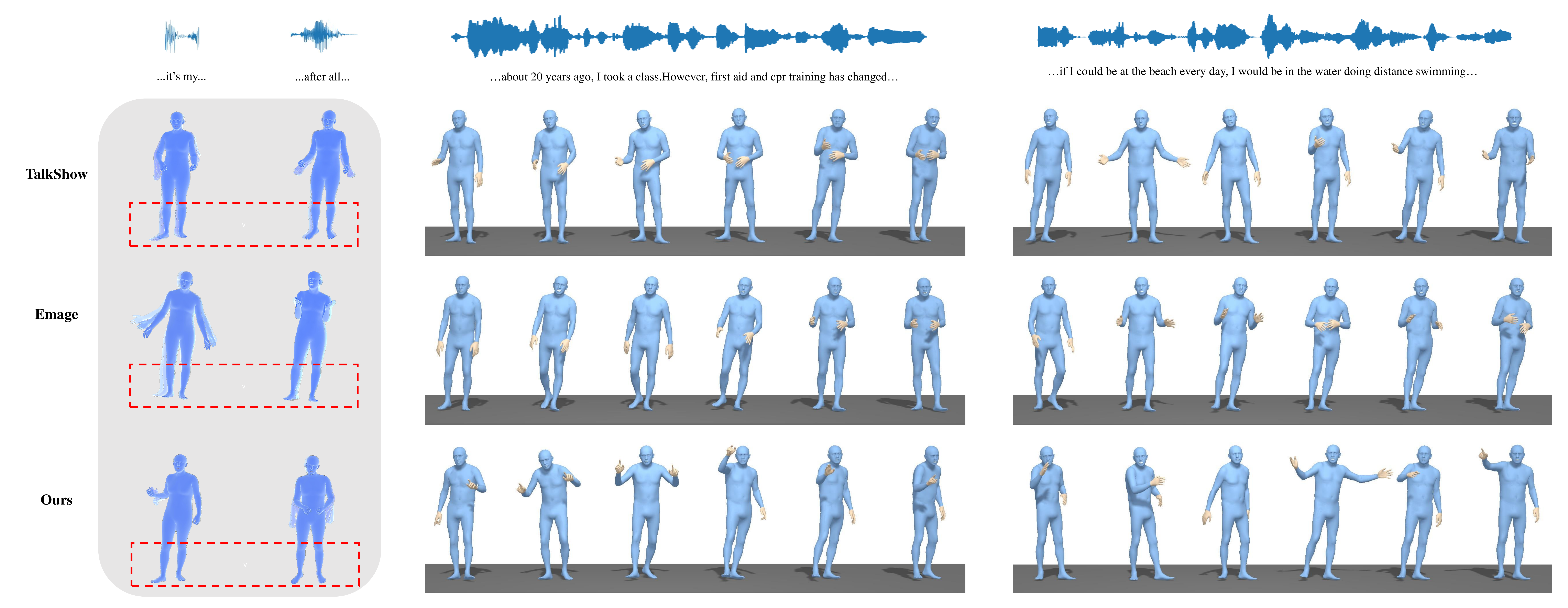}
		\caption{{Qualitative results compared with TalkShow \cite{yi2023generating} and EMAGE \cite{liu2023emage}. The left subfigure shows the continuity and the smoothness of the generated motions, and the right subfigure presents the diversity of the results. In the left subfigure, the first row shows the two different audio inputs, the second row presents the related text, and the other rows show the generated results by different methods. Each sample consists of five frames extracted at intervals of 2/15 seconds from a generated motion clip. Lighter colors represent past frames. }}
		\label{fig_res}
	\end{figure*}

 \begin{figure}[!t]
		\centering
		\includegraphics[width=0.9\linewidth]{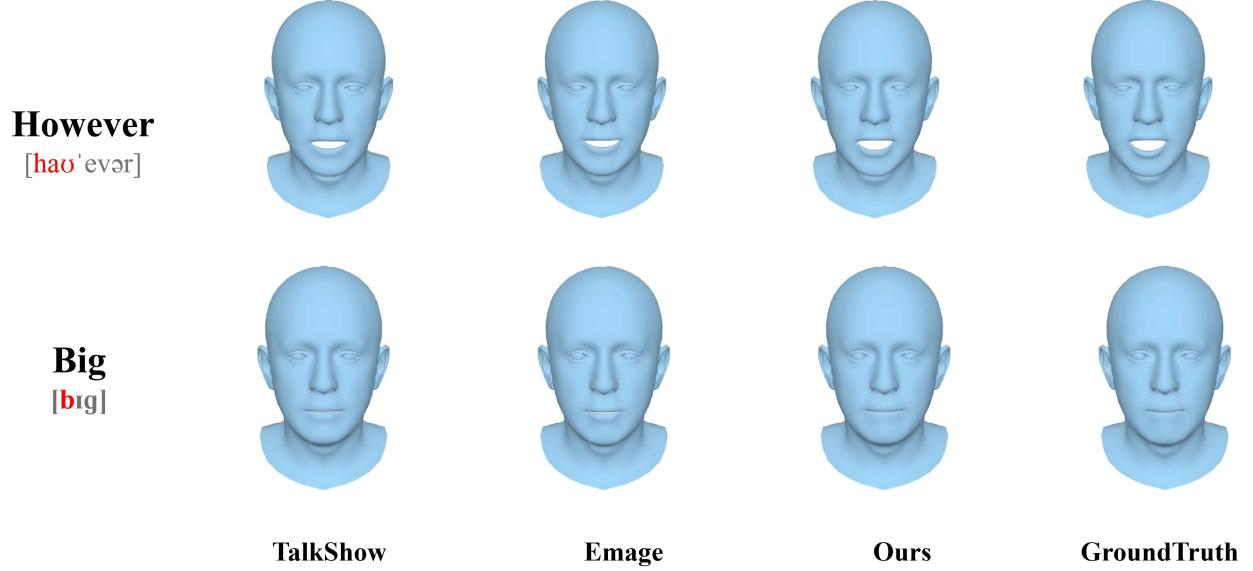}
		\caption{{Comparison of generated facial movements. }}
		\label{fig_face}
	\end{figure}

Besides, to evaluate the deterministic results, we use two metrics to measure the quality of the generated results.

\begin{itemize} 

\item \textbf{LVD: } Landmark Velocity Difference
 evaluates the speed difference between the ground-truth and generated facial landmarks, assessing the alignment between the spoken input and the corresponding face motion.
 
\item
\textbf{LD: } Landmark Distance is used to measure the difference between the ground-truth and generated facial landmarks, including jaw joints and lip shape. In our evaluation, we adopt L2 distance as the measurement.

\end{itemize}

\begin{figure*}[!ht]
		\centering
		\includegraphics[width=\linewidth]{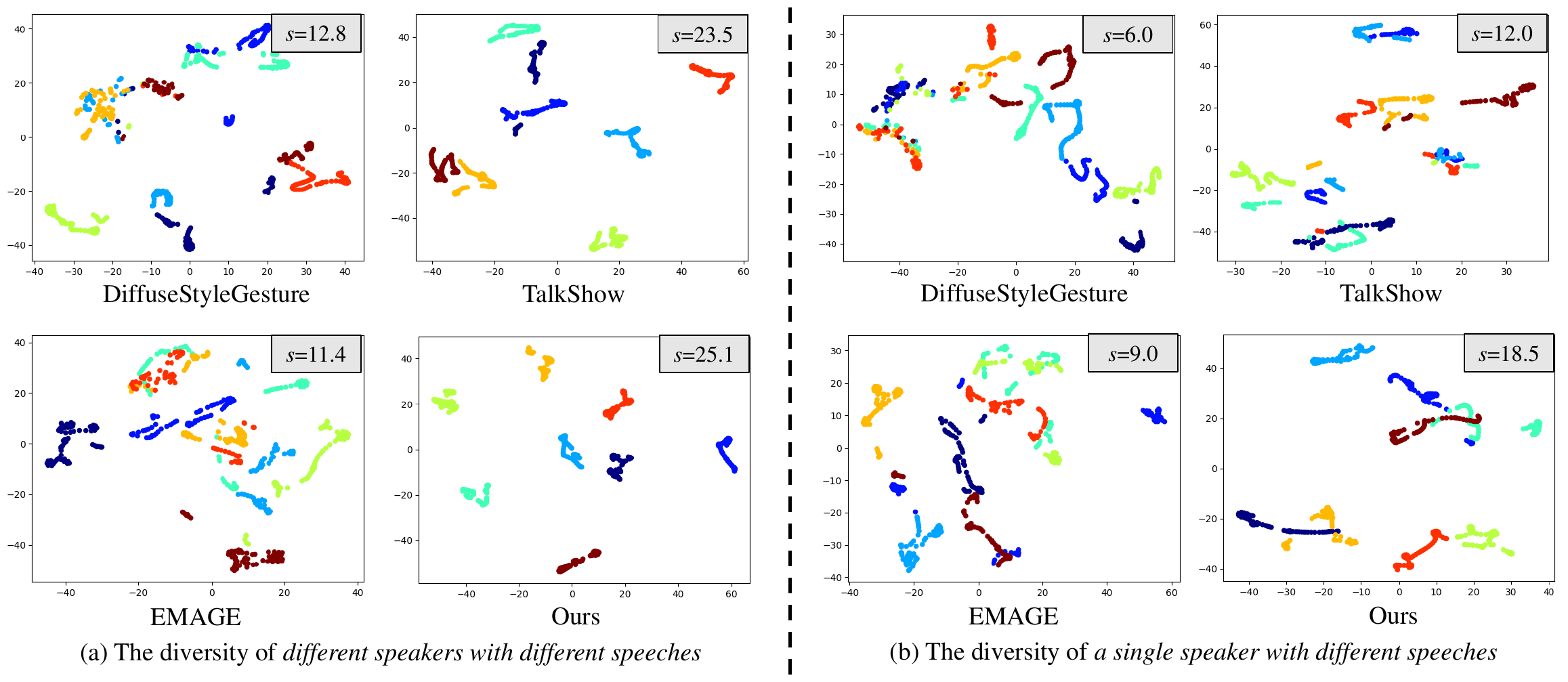}
		\caption{{Visualization of the diversity of different speakers with different {speech audios} and the diversity of a single speaker with different {speech audios} compared with three state-of-the-art methods. The inter-class distance of each method is shown in the top right corner of each figure. Different colors in (a) indicate the visualization of the generated results conditioned on different speakers, and each speaker in the same method has a specific speech audio. Different colors in (b) show the visualization of different speech audios talked by the same speaker.} }
		\label{fig_tsne}
  \vspace{-0.4cm}
	\end{figure*}

\subsection{Comparison Results}

\noindent
\textbf{Quantative Evaluation.}
For body and hand motion generation, Table \ref{tab_com} presents the quantitative comparison on BEAT2 dataset with several baselines.
{It can be seen that our method outperforms other methods in most metrics.}
Specifically, our method significantly improves the performance on FID {(both FID-k and FID-g)}, which indicates that our model generates more realistic body and hand motion sequences.
{The metric FID-k focuses on the speed and acceleration of movement, reflecting the physical realism of the generated motions. Results generated by  previous often contain foot-skating frames and disharmony contacts between ground and feet, which causes unnatural speed and acceleration of the movements. While our method can generate more natural and smooth motion sequences due to our hybrid representation, which leads to the high improvements on FID-K.}
{Besides, the comparison on FSR demonstrates that our model can generate more smooth motions and can alleviate foot skating in generated motions, which suggests the effectiveness of our hybrid representation.}
Also, our method achieves the best performance on BeatAlign score. This denotes that the motion sequences generated by our method are more rhythm-consistent with the input audio.
Moreover, the diversity of our results outperforms other baselines, which indicates our contrastive motion learning method can generate more distinctive motion representations.
{The diffusion-based models, \textit{i.e.}, DiffStyle, requires a large number of diffusion steps, leading to 7 times longer than our method to generate a 4-second motion sequence (5.80s vs. 0.78s). The detailed comparison on inference time can be found in the supplemental document.}

Table \ref{tab_fcom} shows the quantitative comparison on BEAT2 dataset for face motion generation.
According to the comparison, our face generator can synthesize more accurate results aligned with the input audio. This proves the superior performance of our face generator.

\noindent
\textbf{Qualitative Evaluation.}
{Fig. \ref{fig_res} shows the qualitative comparison compared with TalkShow \cite{yi2023generating} and EMAGE \cite{liu2023emage}. In the left subfigure, each sample consists of five frames extracted at intervals of 2/15 seconds from a generated motion clip.}
It can be seen that the foot movements generated by TalkShow and EMAGE change rapidly in a short time, leading to foot skating.
In contrast, due to our hybrid point representation, our model can generate continuous and smooth motion movements.
Besides, our method generates realistic movements corresponding to the audio input, \emph{e.g.}, hands down when emphasizing ``after all".
{The right subfigure in Fig. \ref{fig_res} presents two generated results given two audio inputs. It can be seen that our method can generate more diverse {body and hand motions} for a single sentence, which benefits from our contrastive motion learning method. }
{Fig. \ref{fig_face} presents the comparison of generated facial movements. The results show that our method can generate more accurate motions for both open-mouth and closed-mouth sounds. More dynamic results and analyses of different methods can be found in the  supplementary material. We also provide some long generation samples (over 3 minutes) in the supplemental video.}

{To boost the diversity of the generated results, we design a contrastive motion learning method according to an intuitive idea that the generated motion of specific audio should be different from the generated motions of other speakers and other audios. To verify the performance of our model, we visualize generated results using tSNE method to present the diversity of different speakers with different {speech audios} and the diversity of a single speaker with different {speech audios}. Fig. \ref{fig_tsne} illustrates the visualization of the generated results. We also give the inter-class distance ($s$) to present the diversity between the different clusters. The inter-class distance refers to the measure of separation between different groups.  Fig. \ref{fig_tsne} (a) is obtained from 8 speakers with different {speech audios}, which indicates that the generated motions should be different. It can be seen that our model has better performance, which can generate more distinctive motions different from other speakers. On the contrary, the other methods struggle to generate distinctive motions, leading to similar motions even among different speakers with different sentences. This is also the reason that the inter-class distance of TalkShow (23.5) is closer to our method (25.1). Fig. \ref{fig_tsne} (b) illustrates the diversity of a single speaker with different {speech audios}. We can see that with different {speech audios}, our method achieves better cluster performance and gets the best inter-class distance, which indicates that our method can generate more diverse results for a single speaker.}

 \begin{figure*}[!t]
		\centering
		\includegraphics[scale=0.8]{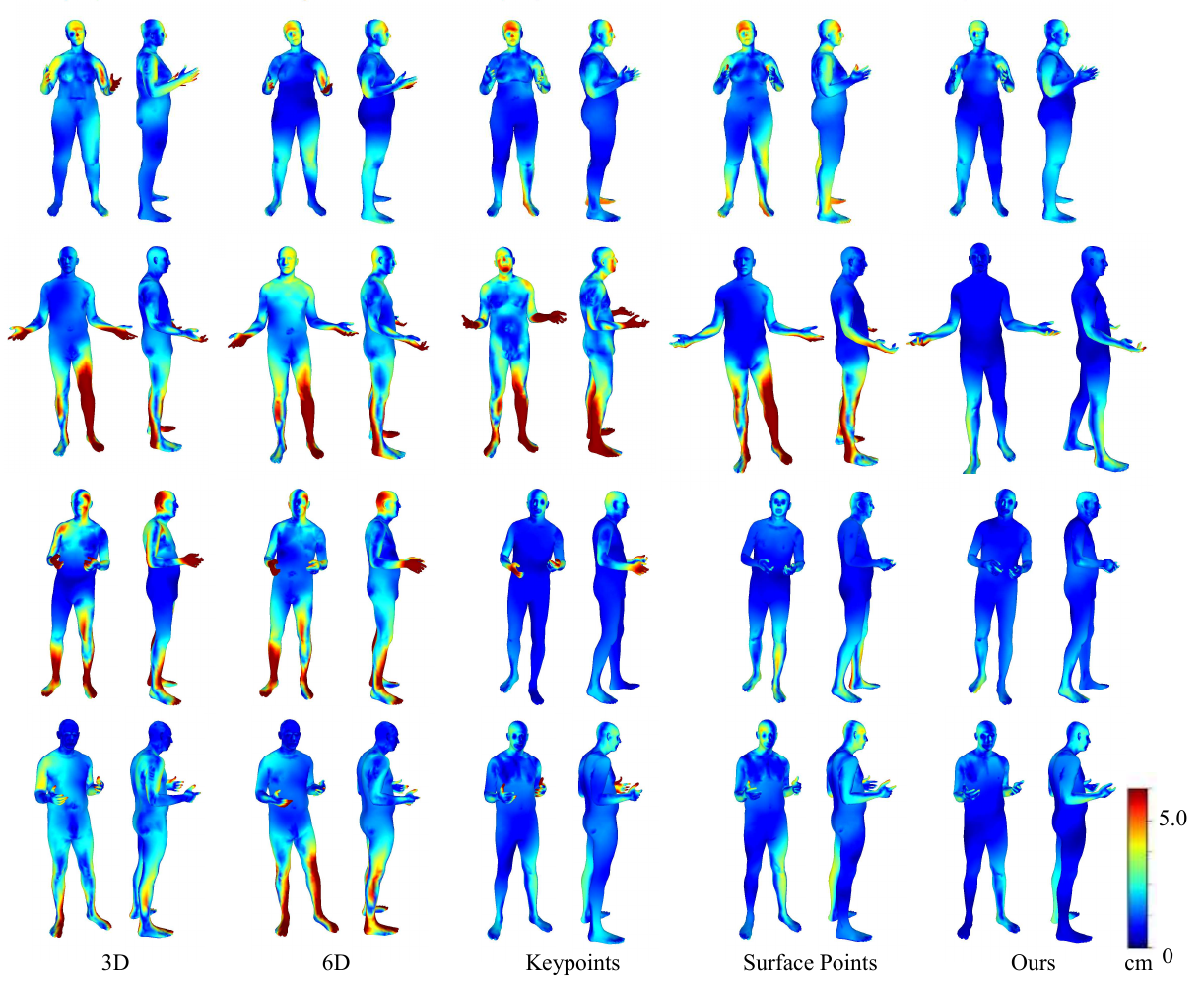}
		\vspace{-0.2cm}
		\caption{Visualization of reconstructed motion errors. The errors between the reconstructed motions and the ground-truths are color-coded on the reconstructed models for visual inspection. {Each row shows reconstructed error for each case.}}
		\label{fig_aba}
  \vspace{-0.2cm}
	\end{figure*}
 
\begin{table}[!t]
	\renewcommand{\arraystretch}{1.4}
	\small%
	\setlength{\tabcolsep}{1.5mm}
	\begin{center}
		\caption{{Ablation studies on BEAT2 dataset for different motion representations. Note that we report LD$\times 10^{2}$ for vertices.}}\label{tab:tab_rec}
  \begin{tabular}{c|cc|cc|c}
			\hline
   \multirow{2}*{Representation} &\multicolumn{2}{c}{Keypoints}&\multicolumn{2}{|c|}{Vertices} &\multirow{2}*{Param.}  \\ \cline{2-5}
			{}&
			    LVD $\downarrow$ & LD $\downarrow$ & LVD $\downarrow$ & LD $\downarrow$ &{} \\
			\hline
			3D (TalkShow)   & 0.4409 & 14.42 & 26.71 & 10.87 & 82.4 M  \\
			6D  &  0.4514 & 15.76 & 26.98 & 11.86 & 82.7 M\\
   {Surface Points}  &  0.3095 & 10.85 & 21.68 & 7.896 & 88.5 M\\
   Keypoints  &  0.2836 & 10.68 & 19.39 & 8.595 & 84.0 M \\
   Ours  &  \textbf{0.2744} & \textbf{7.936} & \textbf{16.68} & \textbf{6.050} & 89.4 M\\
			\hline
		\end{tabular}
\end{center}
\vspace{-0.4cm}
\end{table}

\subsection{Ablation Study}

We conduct extensive experiments to validate the effectiveness of our representation and our model.

\noindent
\textbf{Effectiveness of hybrid point representation.}
To validate the effectiveness of our hybrid point representation, {instead of conducting experiments on the motion generation task, we compare different representations on the motion reconstruction task. Specifically}, we replace our representation with different representations for the first stage of our body generator, \emph{i.e.}, quantized motion modeling (Sec. \ref{vqvae}). 
We employ the 3D representation, 6D representation, surface points of SMPL-X body mesh, and keypoints of SMPL-X body mesh.
{Note that aside from the dimensions of the input and output layers, all other layer configurations, such as the number of layers and hidden dimensions, are identical.}
{ It should be noted that the model with 3D representation has the same representation as TalkShow.}

{We adopt }landmark velocity difference (LVD) and landmark distance (LD) {to evaluate the quality of reconstructed motion}.
To evaluate different representations, we extract keypoints and vertices of reconstructed SMPL-X body meshes to compute LVD and LD metrics.
{In addition, we report the number of parameters (Param.) for different representations. }
Table \ref{tab:tab_rec} presents the quantitative results on BEAT2 dataset for different motion representations. It can be seen that compared with the {surface point} representation, the parametric representation can generate inaccurate results (high LD score) with jitters (high LVD score).
{It is interesting that the performance of 6D representation is lower than 3D representation. The main reason is that speech motion, involves a lot of local rotations, which typically involve small angle changes. Therefore, the 3D axis-angle representation might more easily capture these local characteristics, leading to better performance in the reconstruction task.}
Comparing {surface points} representation and keypoints representation, our hybrid representation can achieve more accurate and smoother results, which indicates that our representation is suitable as a motion representation.
{This is also the reason that our model achieves better results in Table \ref{tab_com} compared with other methods.}

{Fig. \ref{fig_aba} shows the qualitative results. It can be seen that the models based on 3D and 6D representations can produce errors for local body parts, such as hands and legs. The keypoint-based model has better performance, however, the arms are difficult to recover due to a lack of global constraints of the pose rotation. The point-based model can recover more accurate arms, but it struggles to reconstruct the head and legs accurately due to redundant information for some body parts. Compared with these traditional representations, our hybrid point representation can recover the body accurately, which indicates its strong capacity for motion reconstruction. With this powerful representation, we can learn a good motion codebook to further help generate natural motions from speech audios. }

\begin{table}[!t]
	\renewcommand{\arraystretch}{1.3}
	\small%
	\setlength{\tabcolsep}{1.mm}
	\begin{center}
		\caption{Ablation studies on BEAT2 dataset for contrastive motion learning.}\label{tab:tab_con}
		\begin{tabular}{c|ccccc}
			\hline
			{Model} & FID-k $\downarrow$ & FID-g $\downarrow$ & BeatAlign $\uparrow$  & Div-in $\uparrow$ & Div-out $\uparrow$  \\
			\hline
			w/o CM  & 3.865  &  5.972  &   \textbf{0.5767} & 0.8511 & 2.122 \\
			Full & \textbf{3.839} & \textbf{4.359} & 0.5603 & \textbf{0.9700} & \textbf{2.264} \\
			\hline
		\end{tabular}
\end{center}
\vspace{-0.4cm}
\end{table}

\noindent
\textbf{Effectiveness of contrastive motion learning.}
We introduce a novel contrastive motion learning to boost the diversity of generated results.
To validate the effectiveness of this method, we train a body generator without the contrastive motion learning method (w/o CM) as a comparison.
The other settings are the same as our full model.

Table \ref{tab:tab_con} presents the quantitative comparison of BEAT2 dataset.
By comparing the model without the contrastive motion learning method and our full model, it can be seen that the contrastive motion learning method can improve the diversity of generated results.
It is worth noting that the quality-related scores of the model without the contrastive motion learning method still outperform the scores of TalkShow and Audio2Gesture, which suggests the effectiveness of our hybrid point representation.
However, the BeatAlign score of our full model is slightly lower than that of the model without the contrastive motion learning method, which is possibly due to the increase in diversity.

\begin{figure}[!t]
		\centering
		\includegraphics[scale=0.53]{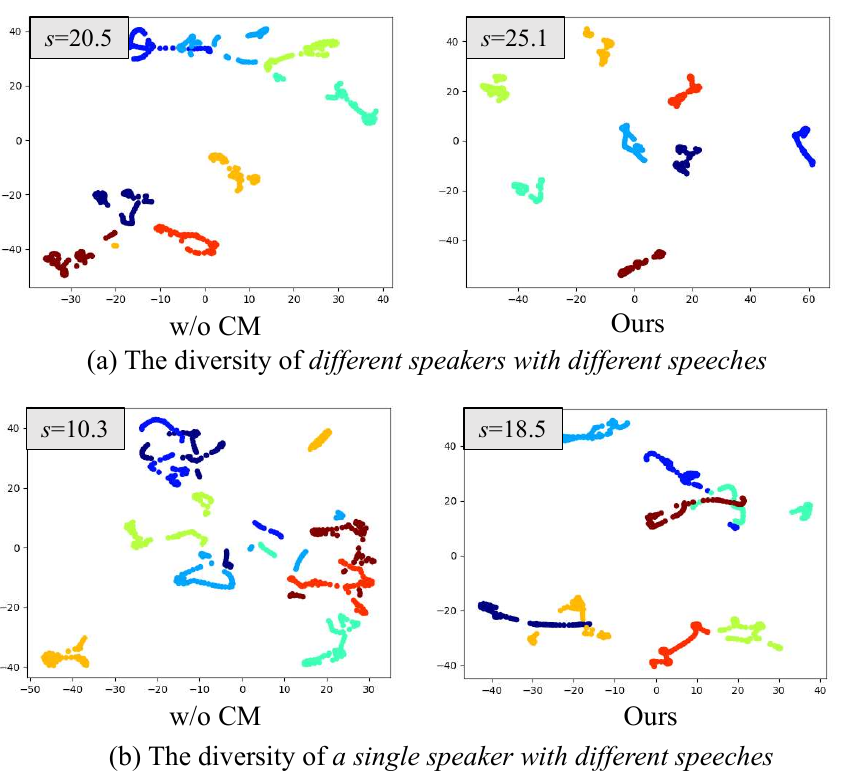}
		\caption{{Visualization of the diversity of different speakers with different {speech audios} (a) and the diversity of a single speaker with different {speech audios} (b). The inter-class distance is shown at the top left corner of each figure.} }
		\label{fig_aba_con}
	\end{figure}

{Fig. \ref{fig_aba_con} shows the visualization of motions generated by our model. Compared with w/o CM and our full model, it can be seen that the contrastive learning method can help the model to generate more distinctive motions for different speakers and different {speech audios}. The increase in the inter-class distances of our full model also verifies this.}

\noindent 
{\textbf{Effectiveness of separate codebooks.}
We use three codebooks, \textit{i.e.,} the body codebook, the left-hand codebook, and the right-hand codebook, to model more detailed motions for the body and hands, respectively. 
To validate the effectiveness of this design, we conduct experiments using different numbers of codebooks.
Specifically, we train motion reconstruction models using one codebook, two codebooks (hand and body codebooks), and three codebooks (our current design).
Table \ref{tab:tab_num} illustrates the quantitative results.
As shown, increasing the number of codebooks improves the performance of motion reconstruction. Although the landmark velocity distance (LVD) for joints is better with two codebooks, the other metrics are lower compared to the model with three codebooks. Therefore, we adopt three separate codebooks in our model.  
}

\begin{table}[t]
	\renewcommand{\arraystretch}{1.4}
	\small%
	\setlength{\tabcolsep}{1.5mm}
	\begin{center}
		\caption{{Ablation studies on BEAT2 dataset for different numbers of codebooks. Note that we report LD$\times 10^{2}$ for vertices.}}\label{tab:tab_num}
  \begin{tabular}{c|cc|cc}
			\hline
   \multirow{2}*{Number} &\multicolumn{2}{c}{Keypoints}&\multicolumn{2}{|c}{Vertices}   \\ \cline{2-5}
			{}&
			    LVD $\downarrow$ & LD $\downarrow$ & LVD $\downarrow$ & LD $\downarrow$  \\
			\hline
		
   1 &  0.2776 & 8.098 & 16.78 & 6.226 \\
   2 &  \textbf{0.2683} & 8.002 & 16.71 & 6.179 \\
   3  &  0.2744 & \textbf{7.936} & \textbf{16.68} & \textbf{6.050} \\
			\hline
		\end{tabular}
\end{center}
\vspace{-0.4cm}
\end{table}

\subsection{User Study}
\label{sec:user}
To better evaluate the proposed method qualitatively, we conduct user studies to analyze the performance of generated motions.
Our questionnaire consists of 10 cases, with each case comprising three questions.
The motions for each case in the video are generated based on the same audio input using Audio2Gesture \cite{li2021audio2gestures}, TalkShow \cite{yi2023generating}, and our method.
 Users are required to rank the results of different methods from the following three perspectives: 
1) the realism of the generated motion; 2) the matching degree between the generated motion and the input audio; and 3) the diversity of the generated motion.
We have collected answers from 153 participants, including 85 males and 68 females with different ages (3 users below 18, 105 users between 18 and 40, 44 users between 40 and 60, and 1 user beyond 60).

\begin{figure}[!t]
		\centering
		\includegraphics[scale=0.54]{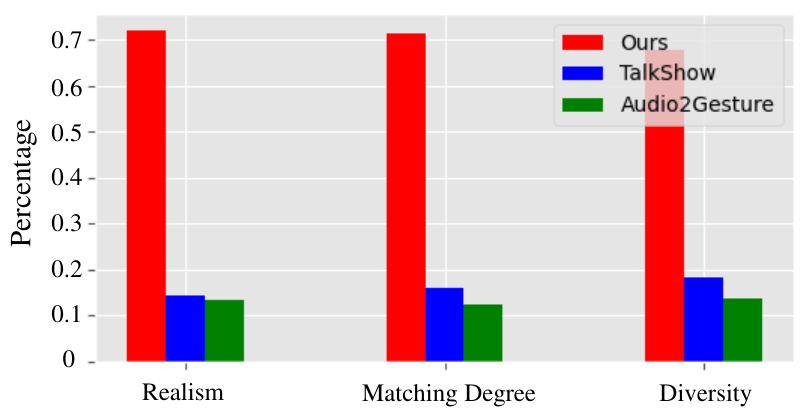}
		\vspace{-0.3cm}
		\caption{The percentage of each method considered to be ranked first in different perspectives. }
		\label{fig_user}
  \vspace{-0.3cm}
	\end{figure}

We evaluate the percentage of each method considered to be ranked first in different perspectives.
Fig. \ref{fig_user} shows the statistical results. 
it is evident that more than 70\% of users believe that our method generates more realistic results, which indicates that our hybrid point representation can produce more smooth and natural results.
Besides, compared to other methods, more than 70\% of users think that our method produces results more in line with the audio rhythm, and 68\% of users believe that our method can generate more diverse motion.
{We also conducted a significance test between our method and the other two methods, which indicates that our method shows statistically significant differences compared to the other two methods. }
These results demonstrate that our method surpasses other methods in terms of diversity and alignment with audio signals, which suggests the effectiveness of our hybrid point representation and our contrastive motion learning method.
{Details of the user study and the significance test can be found in the supplemental document.}

\begin{figure}[!t]
		\centering
		\includegraphics[scale=0.4]{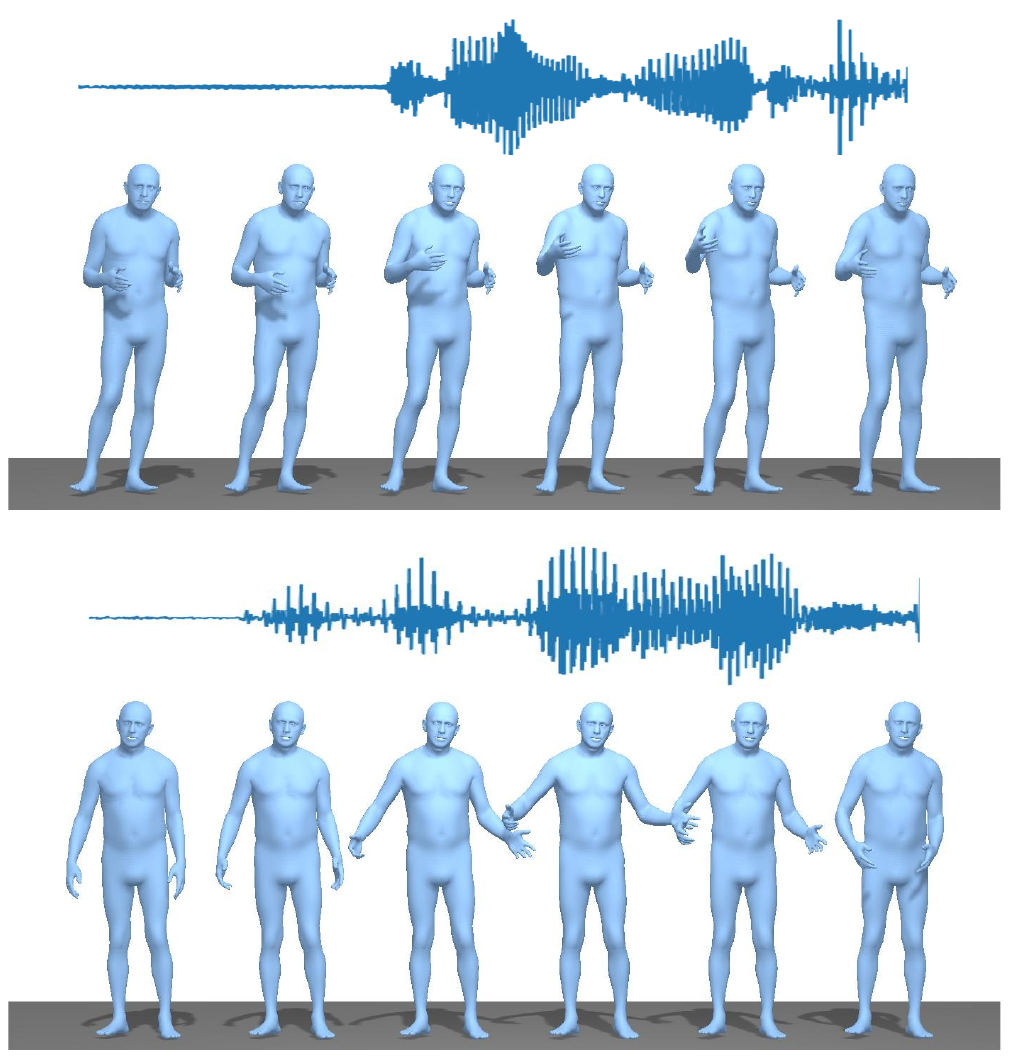}
		\vspace{-0.3cm}
		\caption{{Visualization of generalization across different languages and sentences. The first sequence is in Chinese, and the second sequence is in Spanish.} }
		\label{fig_gen}
   \vspace{-0.3cm}
	\end{figure}


\vspace{-0.3cm}
\subsection{{Generalization}}
{Our model generates natural and diverse motions from only audio signals. Therefore, even with training with English data of BEAT2 dataset \cite{liu2023emage}, our model can be generalized to other languages. }
Fig. \ref{fig_gen} shows the qualitative results. We can see that with the specific rhythm, our model can generate corresponding motions. 


\begin{figure}[!t]
		\centering
		\includegraphics[scale=0.2]{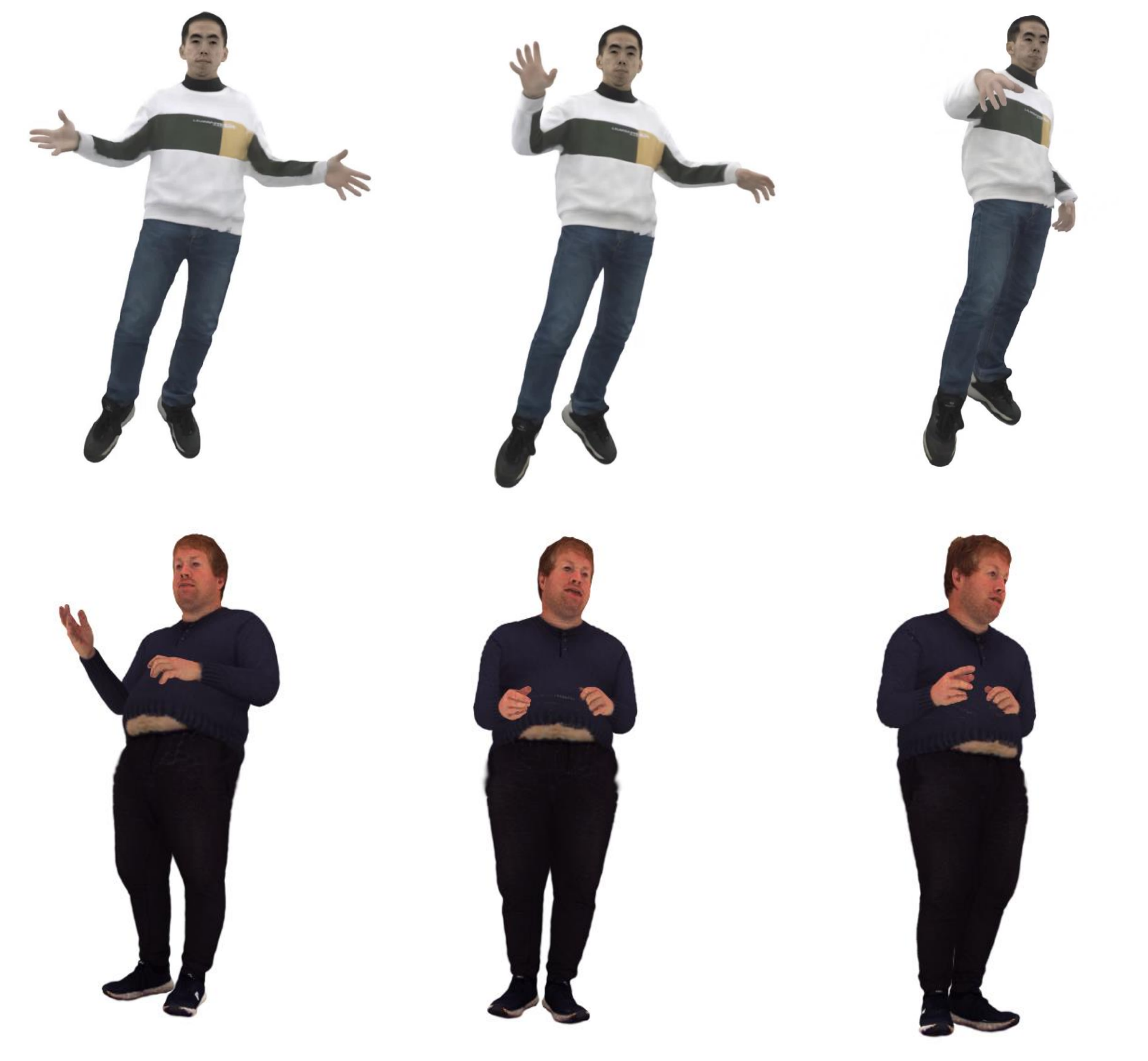}
		\vspace{-0.3cm}
		\caption{{Avatar animation results. Given a {speech audio}, the generated motion can be used to animate reconstructed avatars. } }
		\label{fig_ava}
  \vspace{-0.4cm}
	\end{figure}

 \vspace{-0.4cm}
\subsection{{Application}}
{Thanks to our hybrid point representation, our model can output SMPLX parameters to support many applications. In this section, we obtain an avatar reconstructed from \cite{zheng2023avatarrex}, and animate it using the outputs of our model. By doing so, we can get animated avatars directly using speech, which can be used in VR/AR, virtual live, and human-computer interaction. 
{Since the expression basis of the facial model, \emph{i.e.}, ARKit, used in AvatarRex differs from ours, \emph{i.e.}, SMPLX, we retrained our model using the expression format of ARKit~\cite{linowes2017augmented}. As a result, with the output motions generated by our method, we can directly animate the avatar reconstructed using AvatarRex. To showcase additional results, we adopt the recent avatar reconstruction method ExAvatar~\cite{moon2025expressive} to produce animation results based on our generated motion sequence. Notably, we can directly use the first 50 dimensions of our expression parameters as the expression parameters for ExAvatar, which ensures compatibility with our method without the need for retraining a new model. Fig. \ref{fig_ava} presents the avatar animation results reconstructed from AvatarRex (the first row) and ExAvatar (the second row).}
It can be seen that the generated motion can be applied to avatar animation, and can obtain promising results.   The dynamic results can be found in the {supplementary video.} }

\section{Conclusions and Discussions}

\noindent
\textbf{Conclusion. }
In this paper, we propose a novel framework, named SpeechAct, with a face generator and a body generator for whole-body motion generation from speech audio.
The face generator is used to generate deterministic results using an encode-decoder architecture.
For the body generator,
we propose a hybrid point representation for body and hand to constrain the global surface and capture local details for 3D body, which can achieve accurate yet continuous results.
Based on our representation, we design a two-stage model with a novel contrastive motion learning method to achieve diverse body and hand motions.
Experimental results demonstrate that our model can generate natural and diverse human motion.
{We give some promising results for generalization across other languages. Moreover, an application of speech-driven avatar animation is given to show the potential of VR/AR and human-computer interaction.}

\noindent
\textbf{Limitation and Future Work.}
{To achieve natural results, we estimate the global position using a residual estimation manner. Specifically, we estimate position residuals and add previous residuals to obtain the current position, which can help to generate smooth results. However, for long motion generation, the position error will be added and lead to a sliding effect for some specific cases. {The visualization {and discussion} of some samples can be found in the supplemental document.}  In the future, we will consider more temporal constraints to generate more smooth results.}
{Besides, our model can generate diverse results with distinctive styles. However, we only use the speaker embedding as the style condition, which limits the effectiveness of controllability. {In the future, we will add more control inputs, \emph{e.g.}, text, as conditional inputs, and leverage the speaker information to achieve more controllable human motion synthesis.}}

\vspace{-0.4cm}
\section*{Acknowledgements}
  This work was supported in part by  National Key R$\&$D Program of China (2023YFC3082100), National Natural Science Foundation of China (62122058 and 62171317),  and Science Fund for Distinguished Young Scholars of Tianjin (No. 22JCJQJC00040). We are grateful to Associate Editor and anonymous reviewers for their help in improving this paper.
  
\bibliographystyle{IEEEtran}
\bibliography{template.bib}

\vspace{-1cm}

\begin{IEEEbiography}[{\includegraphics[width=1in,height=1.25in,clip,keepaspectratio]{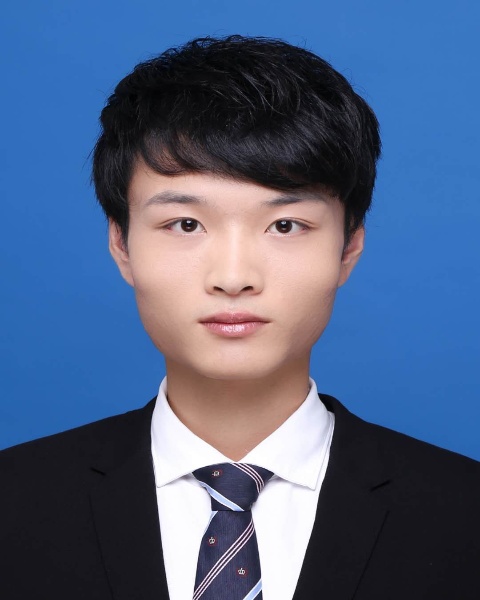}}]{Jinsong Zhang }
received  B.E. and M.E. degrees from Tianjin University, China, in 2018 and 2021, respectively, where he is currently pursuing a Ph.D. degree in computer science. His research interests mainly include  computer vision and computer graphics.
\end{IEEEbiography}

\vspace{-3cm}

\begin{IEEEbiography}
    [{\includegraphics[width=1in,height=1.25in,clip,keepaspectratio]{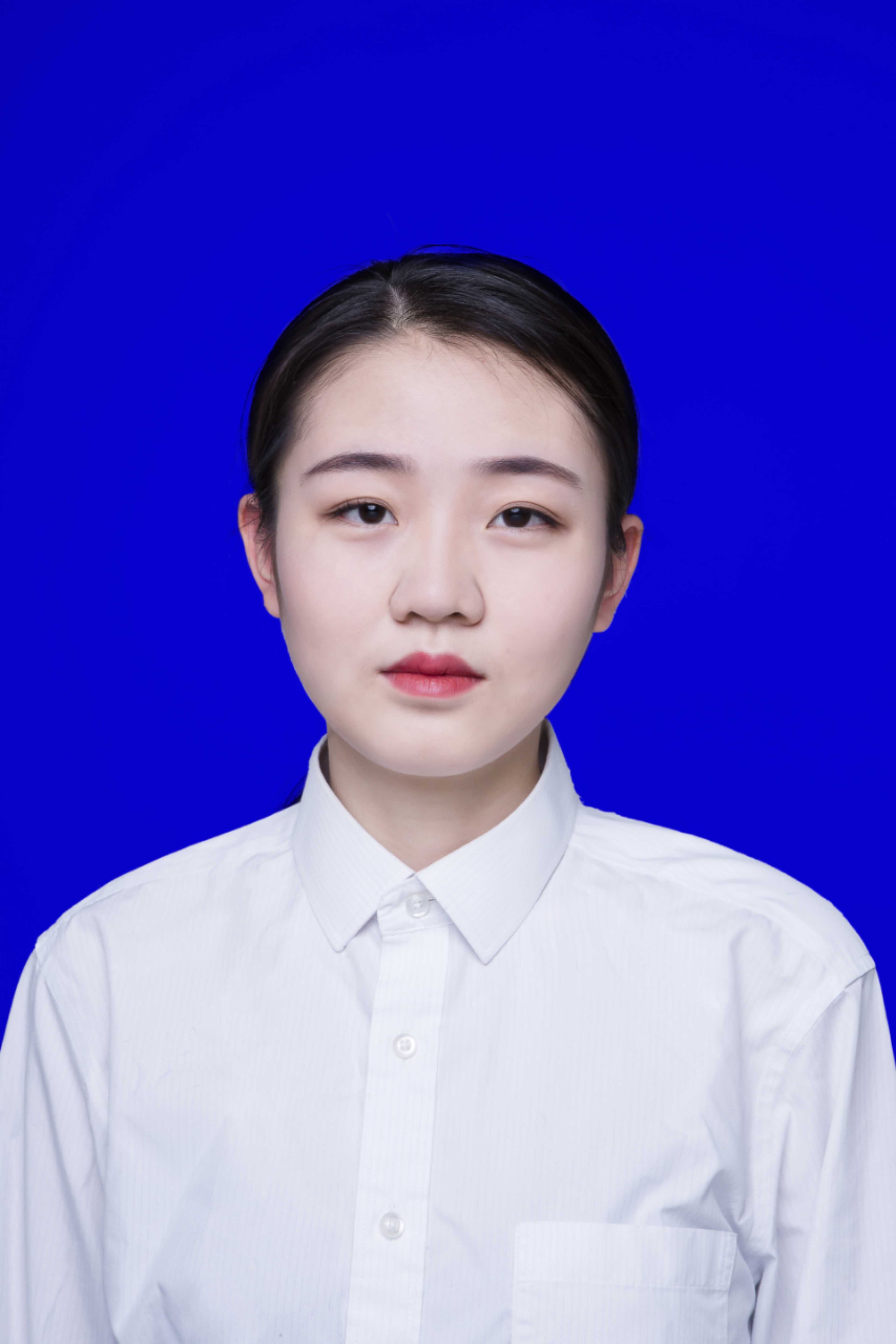}}]{Minjie Zhu }
received her bachelor's degree from Anhui University in 2018 and she is currently studying for master's degree in Tianjin University. Her main research direction is computer vision.
\end{IEEEbiography}

\vspace{-3cm}

\begin{IEEEbiography}
    [{\includegraphics[width=1in,height=1.25in,clip,keepaspectratio]{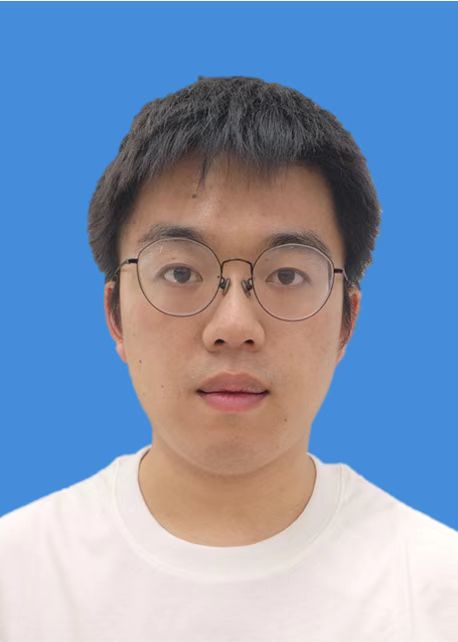}}]{Yuxiang Zhang }
 received the B.E. degree from the Tsinghua University, Beijing, China, in 2019, and the Ph.D. degree from the Tsinghua University, Beijing, China, in 2024. His research interests include 3D vision, human motion capture, 3D human reconstruction and close human interaction recovery.
\end{IEEEbiography}

\vspace{-3cm}

\begin{IEEEbiography}
    [{\includegraphics[width=1in,height=1.25in,clip,keepaspectratio]{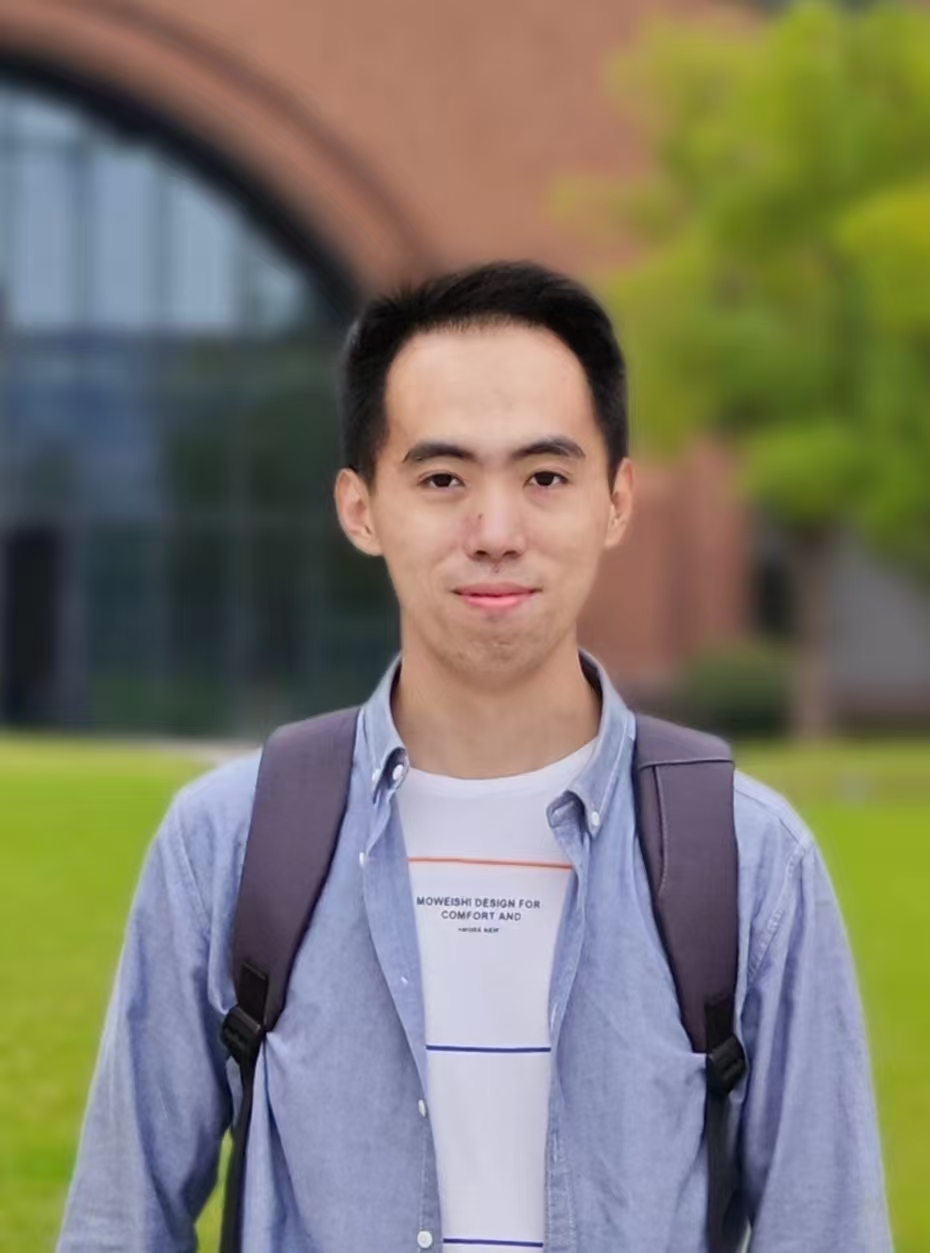}}]{Zerong Zheng }
received the B.S. degree and the PhD degree in Department of Automation, Tsinghua University in July 2018 and July 2023, respectively. He is currently a algorithm scientist in Bytedance. His current research interests include computer vision and computer graphics. 
\end{IEEEbiography}

\vspace{-3.0cm}
\begin{IEEEbiography}
    [{\includegraphics[width=1in,height=1.25in,clip,keepaspectratio]{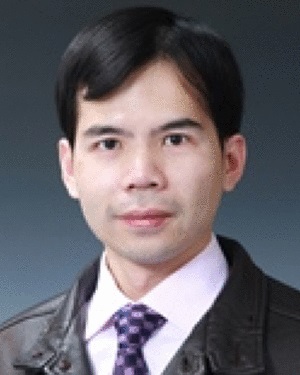}}]{Yebin Liu }
(Member, IEEE) received the B.E. degree from the Beijing University of Posts and Telecommunications, China, in 2002 and the Ph.D. degree from the Automation Department, Tsinghua University, Beijing, China, in 2009. He is currently a full professor with Tsinghua University. He was a re search fellow in the Computer Graphics Group of the MaxPlanck Institute for Information, Germany, in 2010. His research areas include computer vision, computer graphics, and computational photography.
\end{IEEEbiography}

\vspace{-3.0cm}
\begin{IEEEbiography}
    [{\includegraphics[width=1in,height=1.25in,clip,keepaspectratio]{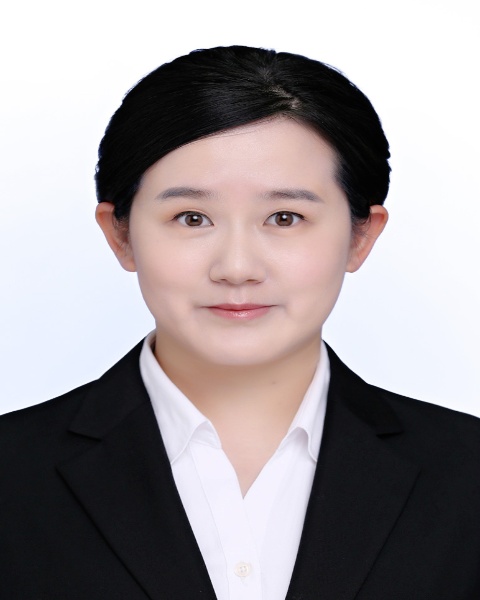}}]{Kun Li }
(Senior Member, IEEE) received the B.E. degree from Beijing University of Posts and Telecommunications, Beijing, China, in 2006, and the master and Ph.D. degrees from Tsinghua University, Beijing, in 2011. She is currently a Professor with the College of Intelligence and Computing, Tianjin University, Tianjin, China. She was the recipient of the CSIG Shi Qingyun Award for Women Scientists. Her research interests include 3D reconstruction and AIGC.

\end{IEEEbiography}

\end{document}